\documentclass[10pt,twocolumn,letterpaper]{article}

\long\def\ignorethis#1{}

\newlength\paramargin
\newlength\figmargin
\newlength\secmargin

\usepackage{array}
\newcolumntype{L}[1]{>{\raggedright\let\newline\\\arraybackslash\hspace{0pt}}m{#1}}
\newcolumntype{C}[1]{>{\centering\let\newline\\\arraybackslash\hspace{0pt}}m{#1}}
\newcolumntype{R}[1]{>{\raggedleft\let\newline\\\arraybackslash\hspace{0pt}}m{#1}}

\newcommand{\secref}[1]{Section~\ref{sec:#1}}
\newcommand{\figref}[1]{Figure~\ref{fig:#1}}
\newcommand{\tabref}[1]{Table~\ref{tab:#1}}

\usepackage[utf8]{inputenc} 
\usepackage[T1]{fontenc}    
\usepackage{url}            
\usepackage{booktabs}       
\usepackage{amsfonts}       
\usepackage{nicefrac}       
\usepackage{microtype}      
\usepackage{cvpr}
\usepackage{times}
\usepackage{epsfig}
\usepackage{graphicx}
\usepackage{amsmath}
\usepackage{amssymb}
\usepackage{adjustbox, color, paralist, enumerate, booktabs, multirow, array, tabu, float, bm}

\usepackage[breaklinks=true,bookmarks=false]{hyperref}

\cvprfinalcopy 

\def\cvprPaperID{5622} 

\ifcvprfinal\fi
\begin{document}

\title{Multi-Scale Boosted Dehazing Network with Dense Feature Fusion}

\author{Hang Dong$^1$~~~~Jinshan Pan$^2$~~~~Lei Xiang$^1$~~~~Zhe Hu$^3$~~~~Xinyi Zhang$^1$
\\ Fei Wang$^1$~~~~Ming-Hsuan Yang$^{4,5}$
\\
$^1$ College of Artificial Intelligence, Xi’an Jiaotong University
\\$^2$ Nanjing University of Science and Technology~~~~$^3$ Hikvision Research America
\\$^4$ University of California, Merced~~~~$^5$ Google Research
\\
}

\maketitle
\thispagestyle{empty}
\linespread{0.98} 

\begin{abstract}
    %
    In this paper, we propose a Multi-Scale Boosted Dehazing Network with Dense Feature Fusion based on the U-Net architecture.
    The proposed method is designed based on two principles, boosting and error feedback, and we show that they are suitable for the dehazing problem.
  By incorporating the Strengthen-Operate-Subtract boosting strategy in the decoder of the proposed model, we develop a simple yet effective boosted decoder to progressively restore the haze-free image.
    %
  To address the issue of preserving spatial information in the U-Net architecture, we design a dense feature fusion module using the back-projection feedback scheme.
    We show that the dense feature fusion module can simultaneously remedy the missing spatial information from high-resolution features and exploit the non-adjacent features.
    Extensive evaluations demonstrate that the proposed model performs favorably against the state-of-the-art approaches on the benchmark datasets as well as real-world hazy images.
    The source code and supplementary are available at \href{https://github.com/BookerDeWitt/MSBDN-DFF}{https://github.com/BookerDeWitt/MSBDN-DFF} 
    and \href{https://drive.google.com/file/d/1pAuhHFOV6wV1xNJ6ZAMyLZ_APaRkp45O/view?usp=sharing}{google drive}.
  \end{abstract}

  \vspace{-3mm}
  \section{Introduction}
  \label{sec:intro}
  \vspace{-1mm}
  Hazy images are usually degraded by the turbid medium in the atmosphere during the imaging formation process.
  The goal of image dehazing is to restore a clean scene from a hazy image.
  This problem has received significant attention as images need to be first enhanced before applying
  high-level vision tasks (e.g., scene understanding~\cite{dehazing_app1} and detection~\cite{dehazing_app2}).
  Existing methods~\cite{deep_physical1, MSCNN, DCPDN, AOD, PDN} usually model a hazy image $I$ by:
  \begin{equation}\label{eq:1}
  I(x)=T(x)J(x)+(1-T(x))A,
  \end{equation}
  where $J$ denotes a haze-free scene radiance, $A$ describes the global atmospheric light indicating the intensity of ambient light, $T$ is the transmission map, and $x$ represents the pixel position.
  
  To restore the haze-free scene radiance $J$ from a hazy image $I$,
  data-driven deep learning approaches have been demonstrated to be effective.
  Early approaches first use deep Convolutional Neural Networks (CNNs)
  to estimate transmission maps~\cite{deep_physical1, MSCNN, DCPDN, AOD, deep_physical_unsupervised1, jinshan_dual} and then apply conventional methods (e,g.,~\cite{He_dark}) to estimate atmospheric light.
  %
  However, the estimation of the transmission map or the atmospheric light from a single hazy input is not a trivial task, 
  due to the airlight-albedo ambiguity~\cite{VHaze} and the difficulty of obtaining ground truth data of transmission maps.
  In addition, inaccurate estimation of the transmission map or the atmospheric light would significantly interfere with the clear image restoration.
  %
  %
  To address this problem, several algorithms directly~\cite{DcGAN, GFN, MsPPN, PFFNet, GCANet, cycledehaze, DuRN,griddehazenet} or iteratively~\cite{DeepPriorsDehaze, DPN} estimate clean images based on deep CNNs.
  Nevertheless, these methods mainly adopt generic network architectures (e.g., DenseNet~\cite{MsPPN}, U-Net~\cite{PFFNet}, Dilated Network~\cite{GCANet}, Grid Network~\cite{griddehazenet}),
  %
  which are not well optimized for the image dehazing problem.

  Different from many high-level vision tasks, inverse problems such as the image dehazing problem are highly ill-posed, where small measurement errors usually lead to dramatic changes.
  To solve these ill-posed problems, certain priors~\cite{tr_dehaze1,tr_dehaze3,He_dark, NLD} or careful algorithm designs are needed to make the problem well-posed.
  For a dehazing deep network, simply stacking more layers or using wider layers is inefficient for significant performance gain.
  Thus, it is of great interest and importance to tailor design network models for the dehazing problem.
  
  In this work, we propose a dehazing network following two well-established principles for image restoration problems, i.e., boosting and error feedback mechanisms.
  The boosting strategy~\cite{DiffusionBoosting, TwicingBoosting, SOS} is originally developed for image denoising by progressively refining the intermediate result from the previous iteration, and the error feedback mechanism, especially the back-projection technique~\cite{Irani1991,BiBP,DBPN}, is designed for super-resolution to progressively recover details that are missed in the degradation process.
  We first show that the boosting strategy would facilitate the image dehazing task as well.
  Considering these two principles, we propose a Multi-Scale Boosted Dehazing Network (MSBDN) with the Dense Feature Fusion (DFF) based on the U-Net~\cite{UNet,PFFNet} architecture.
  %
  We interpret the decoder of the network as an image restoration module and thus incorporate the Strengthen-Operate-Subtract~(SOS) boosting strategy~\cite{SOS} in the decoder to progressively restore the haze-free image.
  Due to the downsampling operations in the encoder of the U-Net,
  the spatial information compression may not be effectively retrieved from the decoder of the U-Net.
  To address this issue, we propose a DFF module based on the back-projection technique to effectively fuse features from different levels.
  %
  %
  We show that this module can simultaneously preserve spatial information from high-resolution features and exploit non-adjacent features for image dehazing.
  Extensive evaluations demonstrate that the proposed algorithm performs favorably against state-of-the-art dehazing methods.
  
  
  \vspace{1pt}
  The contributions of this work are summarized as follows:
  \begin{compactitem}
  \item We propose a Multi-Scale Boosted Dehazing Network to incorporate the boosting strategy and the back-projection technique neatly for image dehazing.
  \item We show that the boosting strategy can help image dehazing algorithms under a certain axiom and show that the network design with the boosting strategy is simple but effective in practice.
  \item We demonstrate that the Dense Feature Fusion module based on the back-projection technique can effectively fuse and extract features from different scales for image dehazing, and help improve the performance of dehazing networks.
  \end{compactitem}
  %
  
  \vspace{-1mm}
  \section{Related Work}
  \label{sec:related}
  \vspace{-1mm}
  \noindent{\bf Image dehazing.}
  Since image dehazing is an ill-posed problem, existing methods often use strong priors or assumptions as additional constraints to restore the transmission map, global atmospheric light, and scene radiance~\cite{tr_dehaze1,tr_dehaze2,tr_dehaze3,He_dark, NLD}.
  In \cite{tr_dehaze1}, Fattal uses the surface shading information to estimate transmission maps.
  By assuming that haze-free images should have higher contrast than hazy images, a method that enhances the visibility of hazy images by maximizing the local contrast is developed \cite{tr_dehaze2}.
  In \cite{He_dark}, He et al. propose a dark channel prior on the pixel intensities of clean outdoor images and develop a dehazing method using the prior.
  As pixels in a given RGB space cluster are often non-local, Berman et al.~\cite{NLD} develop an effective non-local path prior for image dehazing.
  Since these priors and assumptions are introduced for specific scenes or atmospheric conditions, these dehazing methods are less effective on the scenes when the priors do not hold.
  For example, the dark channel prior~\cite{He_dark} does not perform well for images without zero-intensity pixels.

  To address these problems, numerous data-driven methods based on deep learning have been
  developed~\cite{deep_physical1, MSCNN, DCPDN, AOD, PDN, li2018image,DeepPriorsDehaze,deep_physical_unsupervised1,D-DIP} to first estimate transmission maps and then restore images.
  These algorithms are effective when the transmission maps and atmospheric lights are accurately estimated.
  However, due to the airlight-albedo ambiguity~\cite{VHaze}, they usually lead to results with significant color distortions when the estimated atmospheric lights or transmission maps are not accurate enough.
  On the other hand, end-to-end~\cite{DcGAN, GFN, MsPPN, PFFNet, GCANet, DuRN, griddehazenet, Pix2pixHaze, RI-GAN}
  dehazing networks have been proposed to directly restore clean radiance scenes without estimating transmission maps and atmospheric lights.
  Nevertheless, these methods are mainly based on some generic network architectures without significant modification, which are inefficient for the image dehazing problem.
  
  \noindent{\bf Boosting algorithms for image restoration.}
  Numerous boosting methods have been developed for image denoising~\cite{TwicingBoosting, DiffusionBoosting, SOS} to progressively refine the result by feeding the enhanced previous estimation as the input.
  Recently, the boosting strategies are incorporated with deep CNNs for object classification~\cite{IBCNN, BoostCNN} and image denoising~\cite{DBF, DBF_PAMI}.
  %
  %
  In this work, we show that the boosting strategy can also be applied to image dehazing and incorporate it into network design for this dehazing task.
  
  \noindent{\bf Multi-scale feature fusion.} Feature fusion has been widely used in the network design for performance gain by exploiting features from different layers.
  Numerous image restoration methods fuse features by using dense connection~\cite{MsPPN}, feature concatenation~\cite{RDN} or weighted element-wise summation~\cite{GCANet, GFN_BMVC}.
  Most existing feature fusion modules reuse features of the same scale from previous layers.
  In \cite{MsDense, FSSD}, the features at different scales are projected and concatenated
  by using the strided convolutional layer.
  Although these methods can merge multiple features from different levels,
  the concatenation scheme is not effective for extracting useful information.
  To share information among adjacent levels,
  the grid architectures~\cite{error_grid, residual_grid} are proposed by interconnecting the features from adjacent levels with convolutional and deconvolutional layers.
  Recently, Liu et al. propose an effective end-to-end trainable grid network~\cite{griddehazenet} for image dehazing.
  However, these methods do not explicitly exploit features from non-adjacent levels and cannot be easily applied to other architectures.
  In~\cite{DBPN}, a deep feedback mechanism for projection errors~\cite{Irani1991,BiBP} is developed to merge features from two levels.
  Different from \cite{DBPN}, we develop a DFF module which fuses features from multiple scales effectively.
  
  \begin{figure*}
     \centering
     \includegraphics[width=0.81\linewidth]{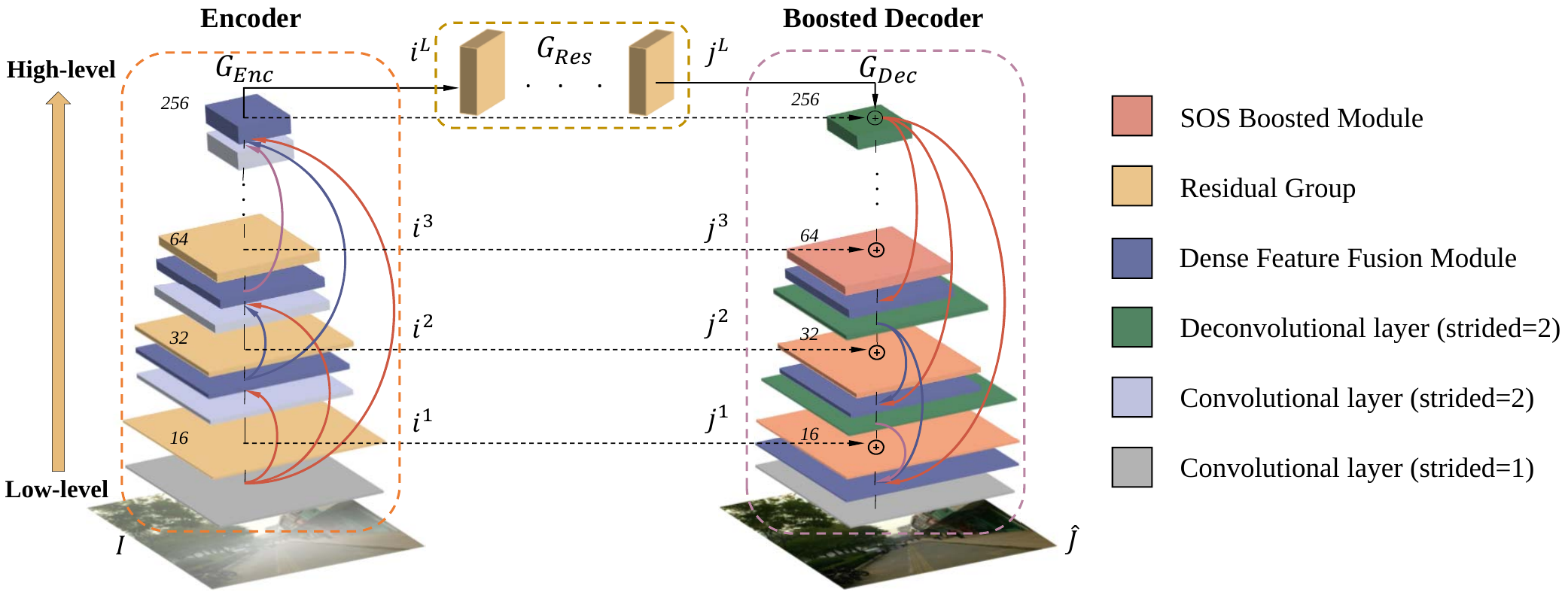}\\
     \caption{\textbf{Architecture of the proposed MSBDN with DFF modules.}
     Skip connections are employed to introduce the corresponding feature maps from the encoder module to the boosted decoder module.}
   \label{fig:1}
   \vspace{-6mm}
   \end{figure*}
  
   \vspace{-2mm}
   \section{Proposed Method}
   \label{sec:3}
  \vspace{-1mm}
  \subsection{Multi-Scale Boosted Dehazing Network}
  \label{sec:3.1}
  \vspace{-1mm}
  The proposed network is based on the U-Net~\cite{UNet} architecture, and we design a multi-scale boosted decoder inspired by the SOS boosting method~\cite{SOS}.
  %
  As shown in \figref{1}, the network includes three components, an encoder module $G_{Enc}$, a boosted decoder module $G_{Dec}$, and a feature restoration module $G_{Res}$.
  %
  
  \noindent{\bf Boosting in image dehazing.}
  The boosting algorithm has been shown to be effective for image denoising \cite{TwicingBoosting, DiffusionBoosting, SOS}.
  The SOS boosting algorithm~\cite{SOS} operates the refinement process on the strengthened image, based on the previously estimated image.
  %
  The algorithm has been shown to improve the Signal-to-Noise Ratio (SNR) under the axiom that the denoising method obtains better results in terms of SNR on the images of the same scene but less noise.
  
  For image dehazing, the SOS boosting strategy can be formulated similarly as
  \begin{equation}
    \label{eqn:3-3}
    \hat{J}^{n+1} = g(I + \hat{J}^{n}) - \hat{J}^{n},
  \end{equation}
  where $\hat{J}^{n}$ denotes the estimated image at the $n$-th iteration, $g(\cdot)$ is the dehazing approach, and $I + \hat{J}^{n}$ represents the strengthened image using the hazy input $I$.
  %
  %
  %
  We show that the boosting method can facilitate image dehazing performance in terms of Portion of Haze (PoH) under a similar axiom as that for denoising.
  Here the portion of haze of the image $J$ in \eqref{eq:1} is defined as $PoH(J) = (1-T)A / J$, and it is proportional to $1 - T$ for hazy images of the same scene.
  
  \vspace{-2mm}
  {\flushleft \bf Axiom 1.}
  \emph{The dehazing method $g$ obtains better results in terms of PoH on the images of the same scene but less haze. That is, if $J_1$ and $J_2$ are the images of the same scene, and $PoH(J_1) < PoH(J_2)$, then $PoH(g(J_1)) < PoH(g(J_2))$.}
  {\flushleft \bf Proposition 1.}
  \emph{Under Axiom 1, the SOS boosting strategy in \eqref{eqn:3-3} improves the dehazing performance, as}
  \begin{equation}
  \label{eqn:3-4}
    PoH(\hat{J}^{n+1}) < PoH(\hat{J}^{n}).
  \end{equation}
  The experimental verification of \textbf{Axiom 1} and proof of \textbf{Proposition 1} are given in the supplementary material.

  According to {\bf Proposition 1}, we develop a deep boosted network based on the SOS boosting strategy, to effectively solve image dehazing by a data-driven approach.
  
  \begin{figure*}[!htb]
    \centering
    \begin{adjustbox}{width=.95\linewidth}
    \begin{tabular}{ccccc}
      \includegraphics[]{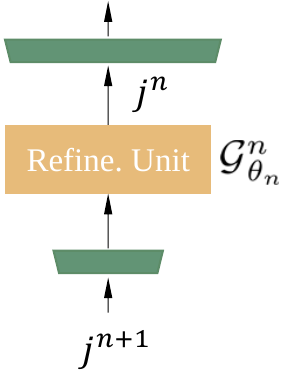} & \hspace{3mm}
      \includegraphics[]{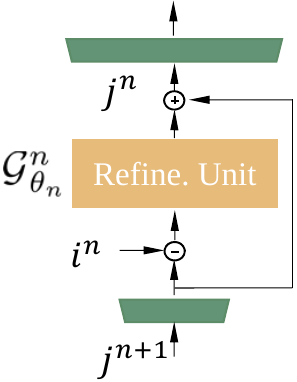} & \hspace{3mm}
      \includegraphics[]{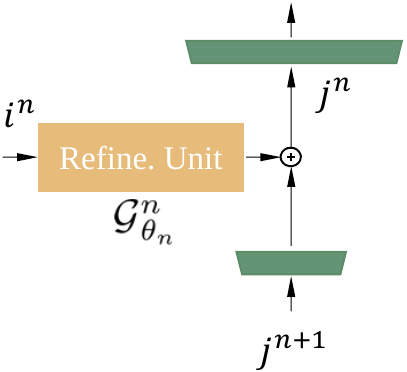} & \hspace{5mm}
      \includegraphics[]{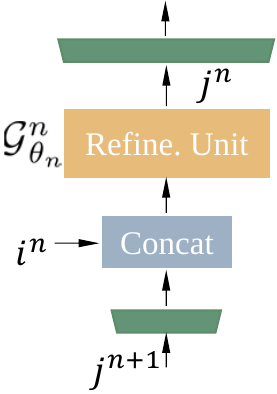} & \hspace{5mm}
      \includegraphics[]{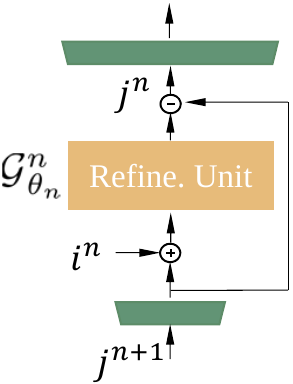} \\
      (a) Diffusion Module \cite{DiffusionBoosting} & (b) Twicing Module \cite{TwicingBoosting} & (c) Pyramid Module~\cite{FPN} & (d) U-Net Module~\cite{UNet} & (e) SOS Boosted Module
    \end{tabular}
  \end{adjustbox}
    \caption{
      \textbf{Architectures of different boosted modules.}
      The green trapezoidal block denotes the deconvolutional layer with a stride of 2 in the decoder module.
    }
    \label{fig:2}
    \vspace{-6mm}
    \end{figure*}
  
  \noindent{\bf Deep boosted dehazing network.}
  In a U-Net network for dehazing, we interpret the decoder as the haze-free image restoration module.
  To progressively refine the feature $j^{L}$ from the feature restoration module $G_{Res}$,
  we incorporate the SOS boosting strategy in the decoder of the proposed network and the structure of the SOS boosted module is illustrated in \figref{2}(e).
  In the SOS boosted module at level $n$, we upsample the feature $j^{n+1}$ from the previous level, strengthen it with the latent feature $i^{n}$ from the encoder,
  and generate the boosted feature $j^{n}$ through the refinement unit, as
  \begin{equation}\label{eqn:3-14}
    j^{n} = \mathcal{G}{}^{n}_{\theta_{n}} (i^{n} + (j^{n+1})\uparrow_{2}) - (j^{n+1})\uparrow_{2},
  \end{equation}
  where $\uparrow_{2}$ denotes the upsampling operator with a scaling factor of 2, $(i^{n} + (j^{n+1})\uparrow_{2})$ represents the strengthened feature,
  and $\mathcal{G}{}^{n}_{\theta_{n}}$ denotes the trainable refinement unit at the $(n)$-th level parameterized by $\theta_{n}$.
  
  In this work, we implement each refinement unit with a residual group as also used in the encoder.
  Clearly, \eqref{eqn:3-14} is derivable and refines the feature $(j^{n+1})\uparrow_{2}$ in a signal strengthening manner.
  %
  %
  At the end of the decoder, a convolutional layer is used for reconstructing the estimated hazy-free image $\hat{J}$ from the final features.
  
  \noindent{\bf Alternatives to SOS boosted module.}
  For completeness, we also list four alternatives to the proposed SOS boosted module for dehazing.
  The diffusion \cite{DiffusionBoosting} and twicing \cite{TwicingBoosting}  schemes are can be applied to designing the boosted modules as shown in \figref{2}(a) and \figref{2}(b).
  They can be formulated respectively as
  \begin{equation}\label{eqn:3-16}
    j^{n} = \mathcal{G}{}^{n}_{\theta_{n}}((j^{n+1})\uparrow_{2}),
  \end{equation}
  and
  \begin{equation}\label{eqn:3-17}
    j^{n} = \mathcal{G}{}^{n}_{\theta_{n}} (i^{n} - (j^{n+1})\uparrow_{2}) + (j^{n+1})\uparrow_{2}.
  \end{equation}
  We adopt the SOS boosting strategy in the proposed method, as the refinement units in \eqref{eqn:3-16} and \eqref{eqn:3-17} do not fully exploit the feature $i^{n}$, which contains more structural and spatial information compared with the upsampled feature $(j^{n+1})\uparrow_{2}$.
  Another related module is the pyramid module (shown in \figref{2}(c)) from the Feature Pyramid Network (FPN) that has been widely used for panoptic segmentation~\cite{PFPN}, super-resolution~\cite{LapSRN}, and pose estimation~\cite{FPN_Pose}.
  It can be formulated as
  \begin{equation}\label{eqn:3-18}
    j^{n} = (j^{n+1})\uparrow_{2} + \mathcal{G}{}^{n}_{\theta_{n}} (i^{n}).
  \end{equation}
  Here, the refinement unit is blind to the upsampled feature $(j^{n+1})\uparrow_{2}$ from the previous level.

  Finally, we also evaluate the decoder module of the original U-Net (shown in \figref{2}(d)), which concatenates the upsampled boosted feature $(j^{n+1})\uparrow_{2}$ and the latent feature $i^{n}$ in the module.
  It can be formulated as
  \begin{equation}\label{eqn:3-19}
    j^{n} = \mathcal{G}{}^{n}_{\theta_{n}}(i^{n}, (j^{n+1})\uparrow_{2}).
  \end{equation}
  Since the subtraction and addition operations in \eqref{eqn:3-14} and \eqref{eqn:3-17} can be absorbed by the learnable refinement unit $\mathcal{G}{}^{n}_{\theta_{n}}$,
  the decoder module of U-Net can imitate the boosting strategy with a proper training.
  However, this imitation is not guaranteed with an implicit and unconstrained fusing process of $(j^{n+1})\uparrow_{2}$ and $i^{n}$.
  
  We evaluate the proposed SOS boosting strategy with these alternatives in Section~\ref{sec:4.2}, and show that the network with the SOS boosting strategy obtains the best results.
  
  \begin{figure*}
    \centering
    \includegraphics[width=0.8\linewidth]{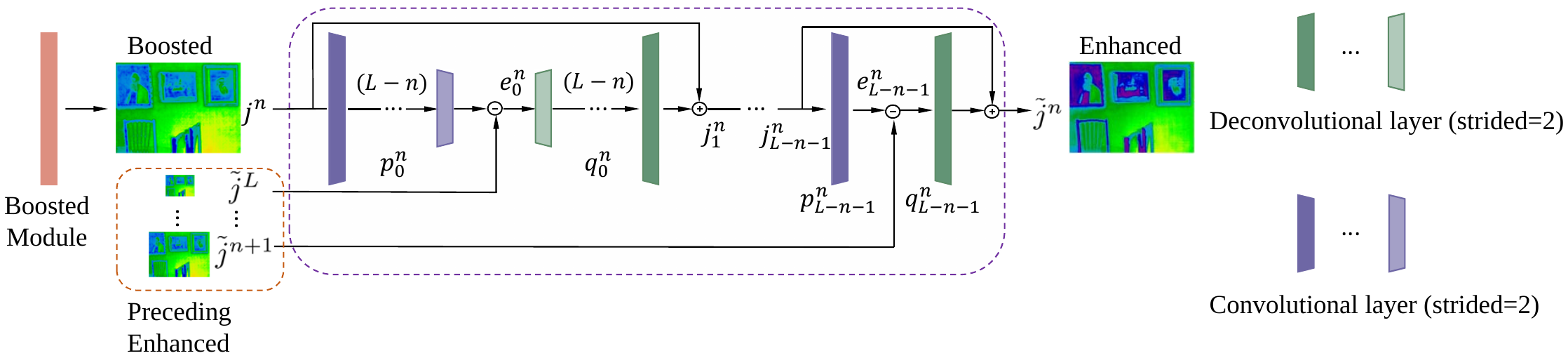}
    \caption{\textbf{Network architecture of the proposed DFF module at the $n$-th level of the decoder.}
  The dense feature fusion in the decoder module uses the back-projection technique to exploit the features from all the preceding levels.
  The blocks with the same color share with the same parameters.}
  \label{fig:3}
  \vspace{-6mm}
  \end{figure*}
  
  \subsection{Dense Feature Fusion Module}
  \label{sec:3.2}
  \vspace{-1mm}
  The U-Net architecture is inherently limited in several aspects, e.g., missing spatial information during the downsampling processes in the encoder and
  lacking sufficient connections between the features from non-adjacent levels.
  To remedy the missing spatial information from upper-level features and fully exploit the features from non-adjacent levels,
  a straightforward approach is to first resample all the features to the same scale,
  and then fuse them by a bottleneck layer (concatenation layer and convolutional layer) as DenseNet~\cite{desnenet} dose.
  However, simply using concatenation is less effective for feature fusion since the features from different levels are of different scales and dimensions.
  
  The back-projection technique in super-resolution~\cite{Irani1991} is an efficient method designed for generating high-resolution contents
  by minimizing the reconstruction errors between the estimated high-resolution result $\hat{H}$ and multiple observed low-resolution inputs.
  In \cite{BiBP}, an iteratively back-projection algorithm is developed for the case with one single low-resolution input with
  \begin{equation}\label{eq:3-9-1}
    \hat{H}_{t+1} = \hat{H}_{t} + h(f(\hat{H_{t}})-  L_{ob}),
  \end{equation}
  where $\hat{H}_{t}$ is the estimated high-resolution output at $t$-th iteration,
  $L_{ob}$ is the observed low-resolution image acquired by using the downsampling operator $f$,
  and $h$ denotes the back-projection operator.
  %
  
  Motivated by the back-projection algorithm in \eqref{eq:3-9-1},
  we propose a DFF module for effectively remedying the missing information and exploiting features from non-adjacent levels.
  The proposed DFF is designed to further enhance the boosted features at the current level
  with an error feedback mechanism, and it is used in both the encoder and the decoder.
  As shown in~\figref{1}, two DFF modules are introduced at each level, one before the residual group in the encoder and another after the SOS boosted module in the decoder.
  The enhanced DFF output in the encoder/decoder is directly connected to all the following DFF modules in the encoder/decoder for feature fusion.
  
  We describe how to use the DFF in the decoder in the following, and the DFF in the encoder can be derived accordingly.
  The DFF for the $n$-th level of the decoder ($\mathcal{D}{}^{n}_{de}$ as shown in \figref{3}),
  is defined by
  \begin{equation}\label{eq:3-10}
    \Tilde{j}{}^{n} = \mathcal{D}{}^{n}_{de}(j^{n}, \{\Tilde{j}^{L}, \Tilde{j}^{L-1}, \cdots, \Tilde{j}^{n+1}\}),
  \end{equation}
  where $j^{n}$ is the boosted feature at the $n$-th level of the decoder,
  $\Tilde{j}{}^{n}$ is the enhanced feature through feature fusion,
  $L$ is the number of the network levels,
  and $\{\Tilde{j}^{L}, \Tilde{j}^{L-1}, \cdots, \Tilde{j}^{n+1}\}$ are the preceding enhanced features from all the $(L-n)$ preceding DFF modules in the decoder.
  Here, we adopt a progressive process to enhance the boosted feature $j^{n}$ by giving one preceding enhanced feature $\Tilde{j}^{L-t}{}, t \in \{0, 1, \cdots, L-n-1\}$ at a time.
  The update procedure can be defined by:
  \begin{compactitem}
    \item Compute the difference $e^{n}_{t}$ between the boosted feature at the $t$-th iteration $j{}^{n}_{t}$ ($j{}^{n}_{0} = j^{n}$) and the $t$-th preceding enhanced feature $\Tilde{j}^{L-t}$ by
     \begin{equation}\label{eq:3-11}
      e^{n}_{t} = p^{n}_{t}(j{}^{n}_{t}) - \Tilde{j}^{L-t},
    \end{equation}
    where $p^{n}_{t}$ denotes the projection operator which downsamples the boosted feature $j{}^{n}_{t}$ to the same dimension of $\Tilde{j}^{L-t}$.
    \item Update the boosted feature $j{}^{n}_{t}$ with the back-projected difference by
    \vspace{-2mm}
    \begin{equation}\label{eq:3-12}
      j{}^{n}_{t+1} = q^{n}_{t}(e^{n}_{t}) + j{}^{n}_{t},
       \vspace{-2mm}
    \end{equation}
    where $q^{n}_{t}$ denotes the back-projection operator which upsamples the difference $e^{n}_{t}$ at the $t$-th iteration to the same dimension of $j{}^{n}_{t}$.
    \item The final enhanced feature $\Tilde{j}^{n}$ can be obtained after all the preceding enhanced features have been considered.
  \end{compactitem}
  Different from the traditional back-projection techniques \cite{Irani1991,BiBP}, the sampling operators $p^{n}_{t}$ and $q^{n}_{t}$ are unknown to the network.
  Motivated by the recent deep back-projection network for super-resolution \cite{DBPN},
  we adopt the strided convolutional/deconvolutional layers to learn the downsampling/upsampling operators in an end-to-end manner.
  To avoid introducing too many parameters, we stack $(L-n-t)$ convolutional/deconvolutional layers with strides of 2 to implement the downsampling/upsampling operators in $p^{n}_{t}$/$q^{n}_{t}$.
  
  We note that the DFF for the $n$-th level of the encoder ($\mathcal{D}{}^{n}_{en}$) can be defined by
   \vspace{-2mm}
  \begin{equation}\label{eq:3-10-1}
    \Tilde{i}{}^{n} = \mathcal{D}{}^{n}_{en}(i^{n}|\Tilde{i}^{1}, \Tilde{i}^{2},\cdots, \Tilde{i}^{n-1}),
     \vspace{-2mm}
  \end{equation}
  where $i^{n}$ is the latent feature at the $n$-th level of the encoder,
  $\{\Tilde{i}^{1}, \Tilde{i}^{2},\cdots, \Tilde{i}^{n-1}\}$ are the preceding enhanced features from all the $(n-1)$ preceding DFF modules in the encoder,
  and $\mathcal{D}{}^{n}_{en}$ shares the same architecture as the module $\mathcal{D}{}^{L-n}_{de}$ at the $(L-n)$-th level of the decoder but switches the positions of downsampling operations $p^{n}_{t}$ and upsampling operations $q^{n}_{t}$.

  Compared with other sampling and concatenation fusion methods, the proposed module can better extract the high-frequency information from the high-resolution features from proceeding layers due to the feedback mechanism.
  By progressively fusing these differences back to the downsampled latent features, the missing spatial information can be remedied.
  On the other hand, this module can exploit all the preceding high-level features and operate as an error-correcting feedback mechanism to refine the boosted features for obtaining better results.
  
  \vspace{-1mm}
  \subsection{Implementations}
  \label{sec:3.3}
  \vspace{-1mm}
  As shown in~\figref{1}, the proposed network contains four strided convolutional layers and four strided deconvolutional layers.
  The Leaky Rectified Linear Unit (LReLU) with a negative slope of 0.2 is used after each convolutional and deconvolutional layer.
  The residual group~\cite{EDSR} consists of three residual blocks, and 18 residual blocks are used in $G_{Res}$.
  The filter size is set as $11 \times 11$ pixels in the first convolutional layer in the encoder module
  and $3 \times 3$ in all the other convolutional and deconvolutional layers.
  We jointly train the MSBDN and DFF module and use the Mean Squared Error~(MSE) as the loss function to constrain the network output and ground truth.
  The entire training process contains 100 epochs optimized by the ADAM solver~\cite{adam} with $\beta_1=0.9$ and $\beta_2=0.999$ with a batch size of 16.
  The initial learning rate is set as $10^{-4}$ with a decay rate of 0.75 after every 10 epochs.
  All the experiments are conducted on an NVIDIA 2080Ti GPU.
  %
  
  \begin{table*}[!t]
    \large
      \centering
      \caption{
        \textbf{Quantitative evaluations on the benchmark dehazing datasets.}
      {\color{red}\textbf{Red texts}} and {\color{blue}blue texts} indicate the best and the second-best performance respectively.
      $\uparrow$ and $\downarrow$ mean the better methods should achieve higher/lower score of this metric.
      }
      \begin{adjustbox}{width=\linewidth}
        \begin{tabu}{lccccccccccccc}
          \tabucline[1.5pt]{}
          \multicolumn{2}{c}{\textbf{Methods}} \hspace{8pt} & DCP~\cite{He_dark}   &NLD~\cite{NLD}  & AODNet~\cite{AOD}   & MSCNN~\cite{MSCNN} & MsPPN~\cite{MsPPN} & DcGAN~\cite{DcGAN}  & GFN~\cite{GFN} & GCANet~\cite{GCANet} & PFFNet~\cite{PFFNet} &GDN~\cite{griddehazenet} &DuRN~\cite{DuRN} & \multicolumn{1}{c}{Ours}  \\
          \hline
          \multirow{2}[2]{*}[2pt]{SOTS~\cite{RESIDE}} & PSNR$\uparrow$  & 18.75 & 17.27 &18.80  & 17.57  &29.94   &25.37  & 24.11 & 28.13 &29.22 &31.51 &{\color{blue}31.92} &{\color{red}\textbf{33.79}} \\
                & SSIM$\uparrow$  & 0.859  &  0.750   &   0.834    &  0.811     & 0.958     &  0.917     &  0.899   &  0.945   &0.954  &{\color{blue}0.983} & 0.980   & {\color{red}\textbf{0.984}}
                \\ \hline
  
          \multirow{2}[2]{*}[2pt]{HazeRD~\cite{HazeRD}} & CIEDE2000$\downarrow$  & 14.83   & 16.40    & 13.23      & 13.80   &  15.50   & {\color{blue}12.02}    & 14.83     & 14.45     & 14.46    & 13.93    & 12.48    & {\color{red}\textbf{10.36}} \\
                                          & SSIM$\uparrow$       & 0.767   & 0.727    & 0.833      & 0.794   &  0.759                               & 0.826    & 0.802     & 0.819     & 0.808    & 0.833   & {\color{blue}0.840}    & {\color{red}\textbf{0.881}}
                 \\ \hline
  
          \multirow{2}[2]{*}[2pt]{I-HAZE~\cite{IHAZE}} & PSNR$\uparrow$  & 14.43 & 14.12 &  13.98& 15.22 & {\color{blue}22.50} &  16.06 &   15.84  &   14.95    &   16.01 &16.62  &21.23 & {\color{red}\textbf{23.93}} \\
                & SSIM$\uparrow$  & 0.752 & 0.654 &  0.732 & 0.755 & {\color{blue}0.871} &       0.733   &   0.751    &   0.719  &   0.740  &0.787 &0.842  &  {\color{red}\textbf{0.891}}
                 \\ \hline
  
          \multirow{2}[2]{*}[2pt]{O-HAZE~\cite{OHAZE}} & PSNR$\uparrow$  & 16.78 & 15.98 & 15.03& 17.56 & {\color{blue}24.24} &     19.34   &  18.16  & 16.28 &18.76  &18.92 &20.45   & {\color{red}\textbf{24.36}} \\
                & SSIM$\uparrow$  & 0.653 &0.585  &0.539 & 0.650 & {\color{blue}0.721} &    0.681  &   0.671   &  0.645  &     0.669  &0.672  &0.688  &  {\color{red}\textbf{0.749}}\\
          \tabucline[1.5pt]{}
        \end{tabu}%
      \end{adjustbox}
      \label{tab:1}%
   \vspace{-4mm}
    \end{table*}
  
  \vspace{-2mm}
  \section{Experimental Results}
  \label{sec:4}
  
  \vspace{-1mm}
  \subsection{Datasets}
  \label{sec:4.0}
  \vspace{-1mm}
  We evaluate the proposed algorithm on the following datasets~\cite{RESIDE,HazeRD,NTIRE2018} against the state-of-the-art methods.
  
  \noindent{\bf RESIDE dataset.}
  The RESIDE dataset~\cite{RESIDE} 
  contains both synthesized and real-world hazy/clean image pairs of indoor and outdoor scenes.
  %
  To learn a general dehazing model for both indoor and outdoor scenes, we select as the training set 9000 outdoor hazy/clean image pairs and 7000 indoor pairs from the RESIDE training dataset~\cite{RESIDE} by removing redundant images from the same scenes.
  To further augment the training data, we resize images of each pair with three random scales within the scale range of $[0.5,1.0]$.
  We randomly crop 256 $\times$ 256 patches from hazy images and randomly flip them horizontally or vertically as the inputs.
  SOTS is the test subset of the RESIDE dataset, which contains 500 indoor hazy images and 500 outdoor hazy images.
  For comparison, all the methods are trained on the selected RESIDE training dataset and evaluated on the SOTS.
  
  \noindent{\bf HazeRD dataset.}
  The HazeRD dataset~\cite{HazeRD} contains 75 synthesized hazy images with realistic haze conditions.
  Since most evaluated methods on the HazeRD dataset~\cite{HazeRD} are trained on the synthesized NYUv2~\cite{NYU} dataset,
  we train the proposed model on NYUv2 with the same setting as \cite{MSCNN} for fair comparisons.
  
  \noindent{\bf NTIRE2018-Dehazing challenge dataset.}
  The NTIRE2018-Dehazing challenge~\cite{NTIRE2018} includes an indoor dataset (referred to as I-HAZE~\cite{IHAZE}) and an outdoor dataset (referred to as O-HAZE~\cite{OHAZE}).
  Both datasets provide training sets and test sets.
  For each dataset, we train the proposed model using its training set and evaluate it on the corresponding test set.
  We adopt the same data augmentation strategy as that used for the RESIDE dataset.
  %
  
  \vspace{-1.5mm}
  \subsection{Performance Evaluation}
  \label{sec:4.1}
  \vspace{-1mm}
  We evaluate the proposed algorithm against state-of-the-art methods based on the hand-crafted priors (DCP~\cite{He_dark} and NLD~\cite{NLD})
  and deep convolutional neural networks (AOD~\cite{AOD}, MSCNN~\cite{MSCNN}, MsPPN\cite{MsPPN}, DcGAN~\cite{DcGAN}, GFN~\cite{GFN}, GCANet~\cite{GCANet}, PFFNet~\cite{PFFNet}, GDN~\cite{griddehazenet}, and DuRN\cite{DuRN}).
  We use the metrics PSNR, SSIM~\cite{SSIM}, and CIEDE2000 \cite{HazeRD} to evaluate the quality of restored images.
  We note that many existing dehazing methods~\cite{GFN,GCANet,PFFNet,DuRN} report their results only on the SOTS indoor images with the models trained on various datasets~\cite{RESIDE, NYU, Middlebury}.
  %
  Moreover, the GDN method~\cite{griddehazenet} reports the results on the SOTS indoor and outdoor sets, with the models trained on the indoor scenes and outdoor scenes separately.
  For fair comparisons, we retrain these methods (GFN, PFFNet, GCANet, MsPPN, and DuRN) using their provided training codes on the same training dataset and evaluate them on the same test dataset, as the proposed algorithm.
  Other methods are evaluated with the provided pre-trained models.
  
  The first row in \tabref{1} shows the quantitative results on the SOTS dataset.
  As expected, the methods based on hand-crafted features~\cite{He_dark, NLD} do not perform well.
  The methods~\cite{AOD, MSCNN, DcGAN} that use deep CNNs in a two-stage restoration (estimating the transmission maps and atmospheric lights first and then hazy-free images), are less effective when the atmospheric lights are not correctly estimated.
  %
  Since the GFN~\cite{GFN} method applies the hand-crafted derived images as inputs, it is less effective on the scenes when these derived images cannot enhance the hazy images.
  On the other hand, the methods~\cite{GCANet,PFFNet, MsPPN, DuRN, griddehazenet} that directly estimate clear images based on end-to-end trainable networks, generate better results than the other indirect ones.
  The proposed method outperforms these algorithms in both PSNR and SSIM, since these networks are not well optimized for the dehazing problem.
  In contrast, the proposed architecture based on the boosting and back-projection technique is more effective for the task.

  The evaluation results on the HazeRD dataset also demonstrate the effectiveness of the proposed algorithm for dehazing realistic images.
  The proposed algorithm achieves better CIEDE2000 and SSIM results than other methods.
  On the NTIRE18-Dehazing challenge~\cite{NTIRE2018}, our method achieves comparable results with the MsPPN~\cite{MsPPN},
  which is specially designed for ultra high-resolution datasets including extra pre-processing and post-processing steps.

\begin{figure*}[!t]
    \centering
    \LARGE
  
    \begin{adjustbox}{width=\linewidth}
      \begin{tabular}{cccccccc}
        \includegraphics[width=0.245\linewidth]{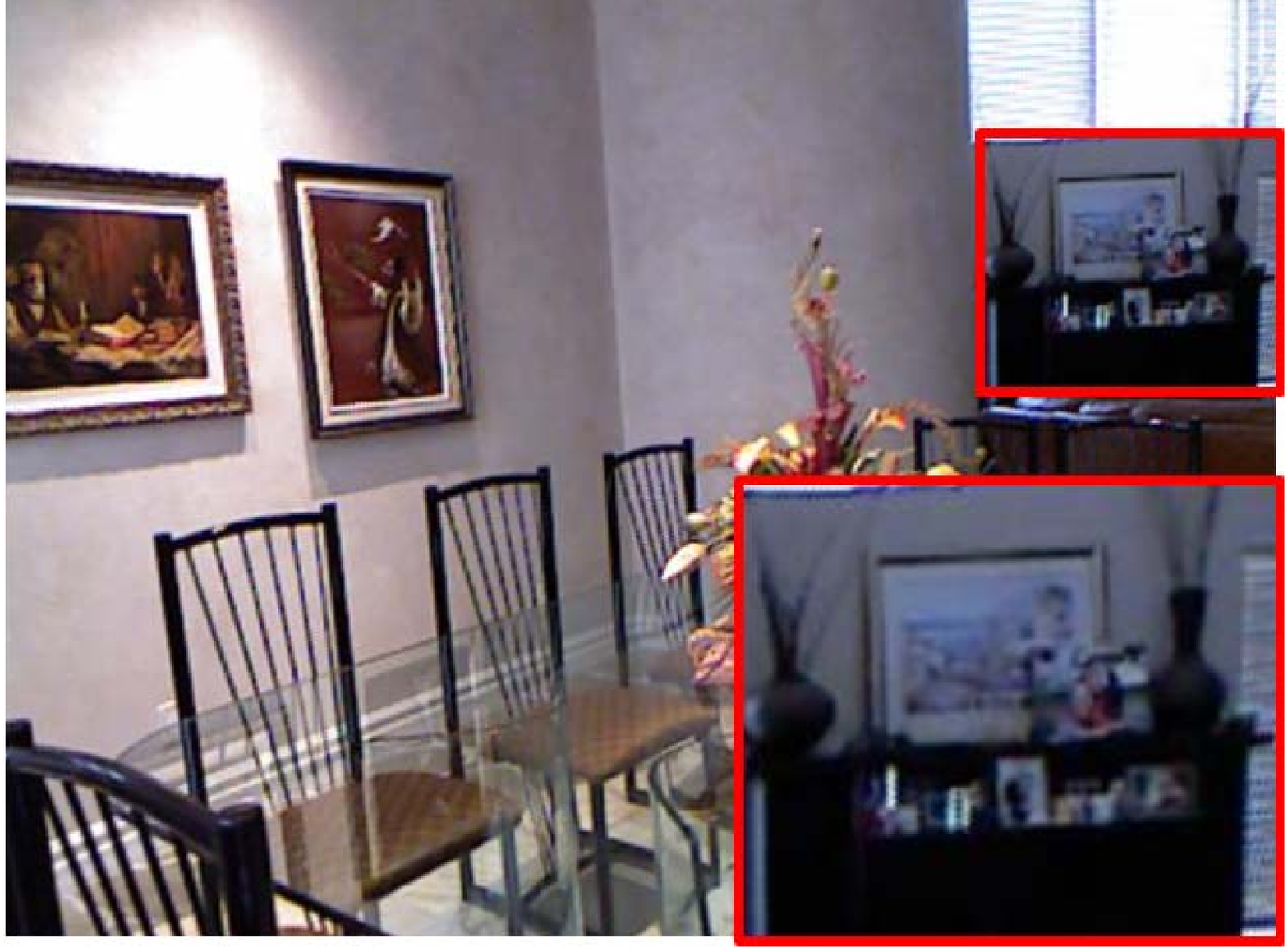} & \hspace{-4mm}
        \includegraphics[width=0.245\linewidth]{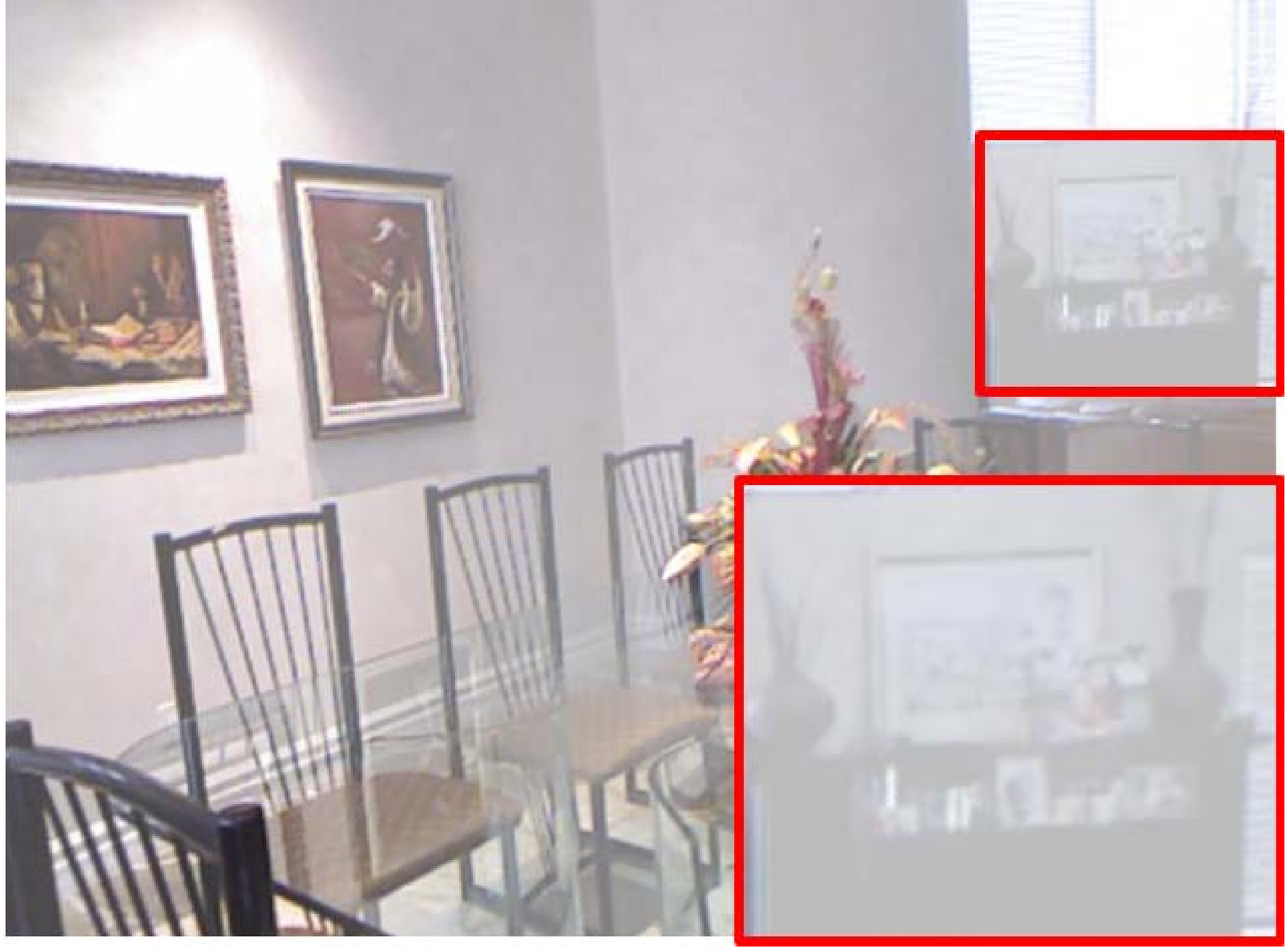} & \hspace{-4mm}
        \includegraphics[width=0.245\linewidth]{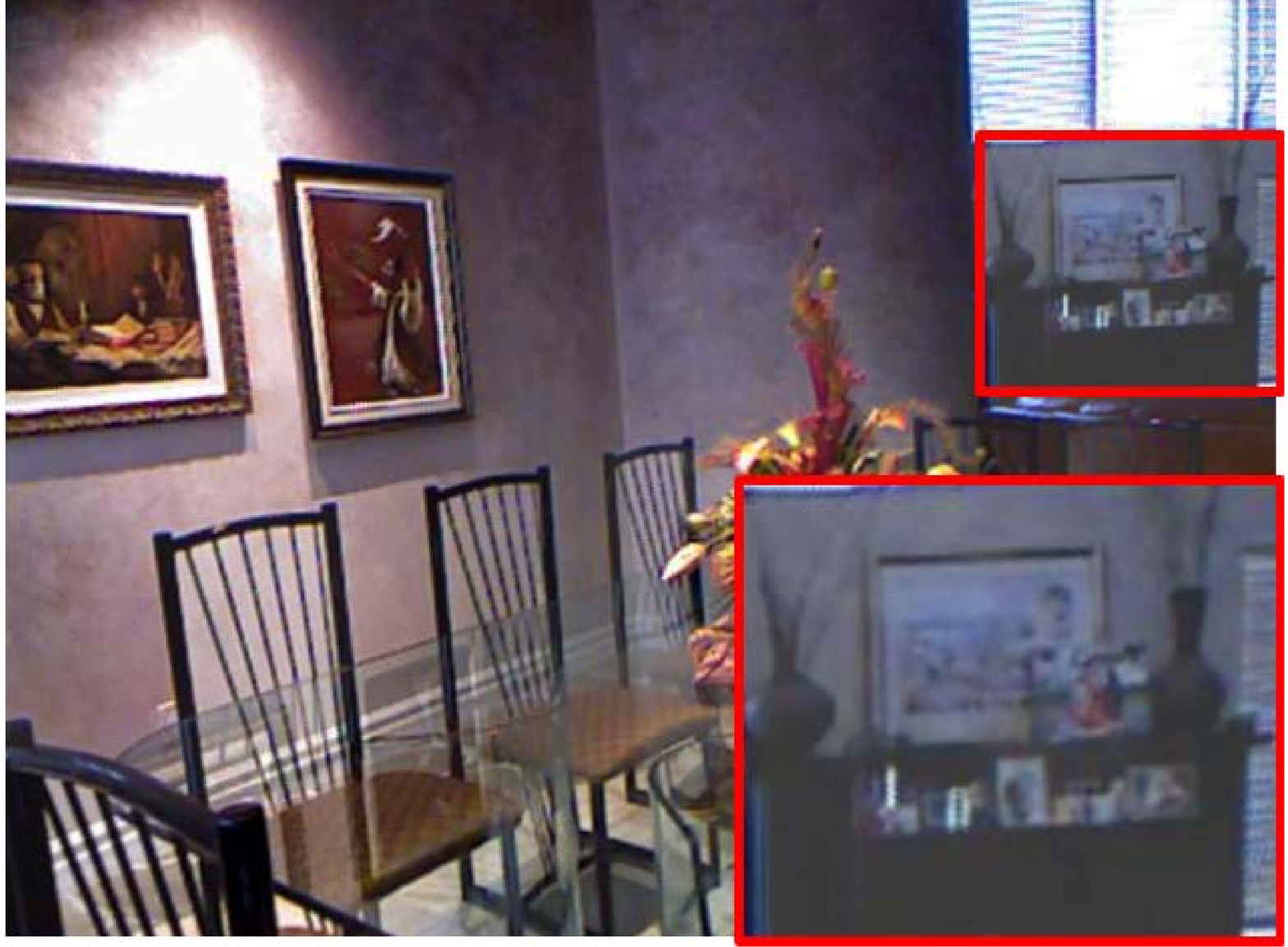} & \hspace{-4mm}
        \includegraphics[width=0.245\linewidth]{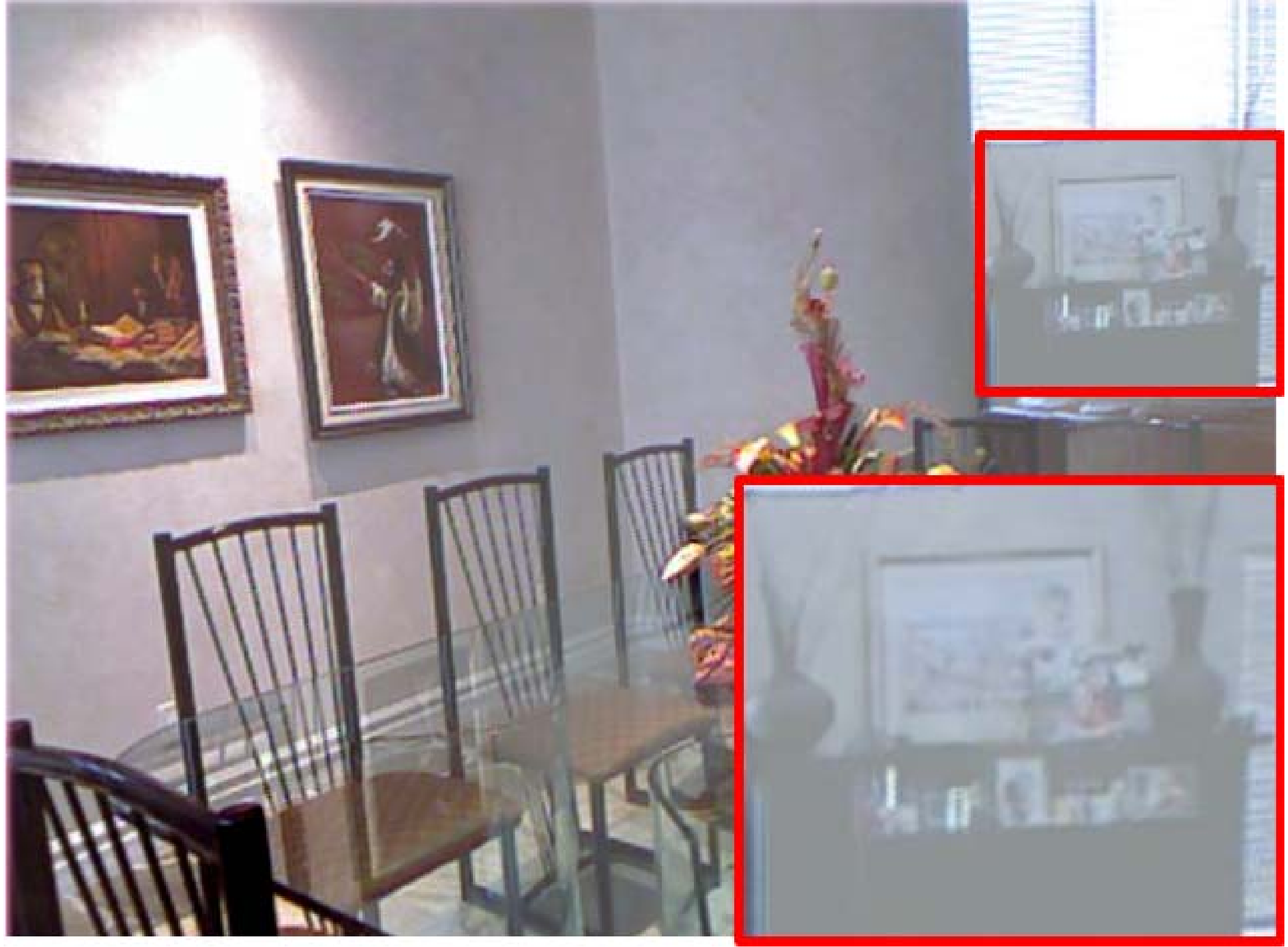} & \hspace{-4mm}
        \includegraphics[width=0.245\linewidth]{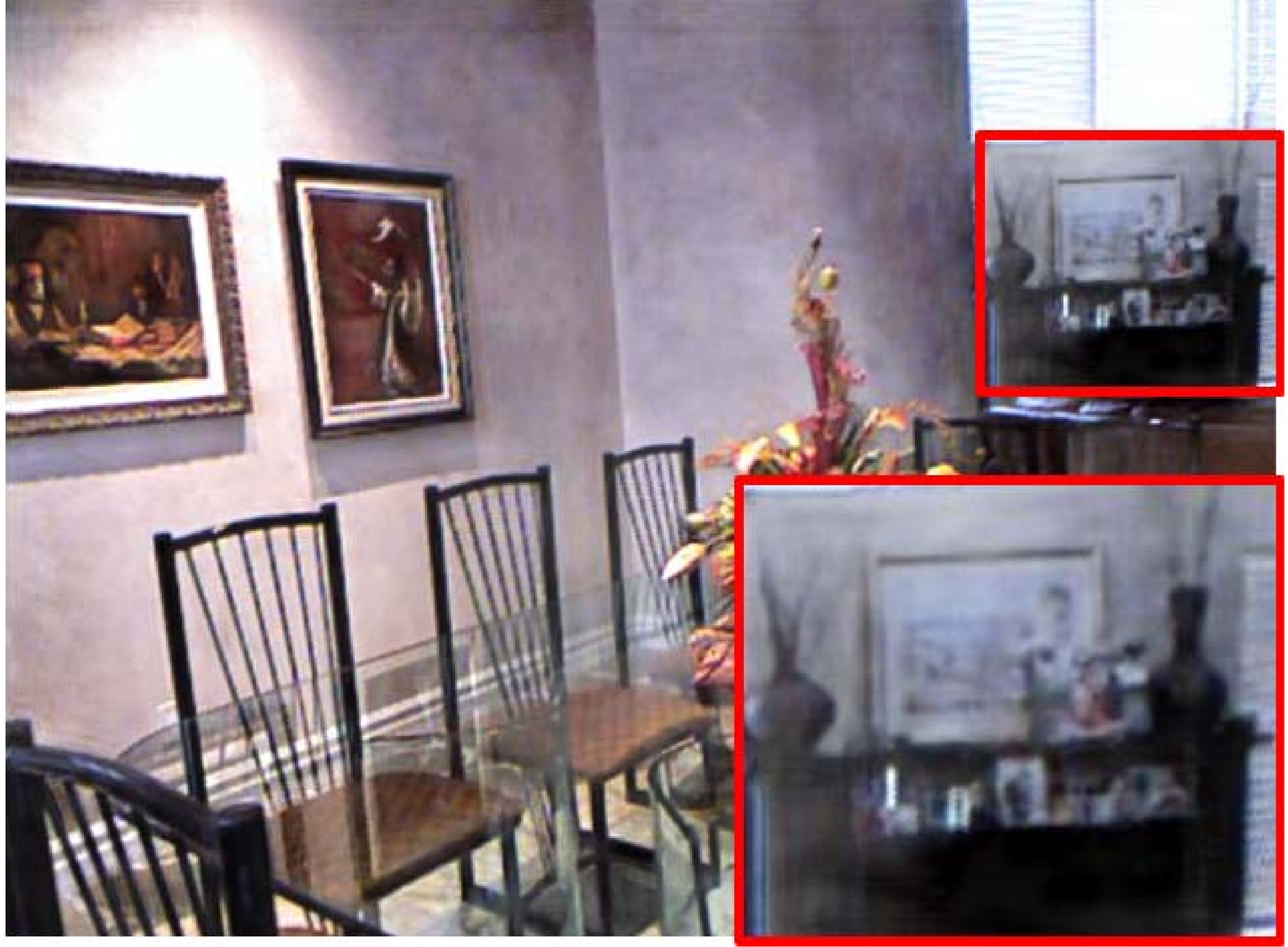} & \hspace{-4mm}
        \includegraphics[width=0.245\linewidth]{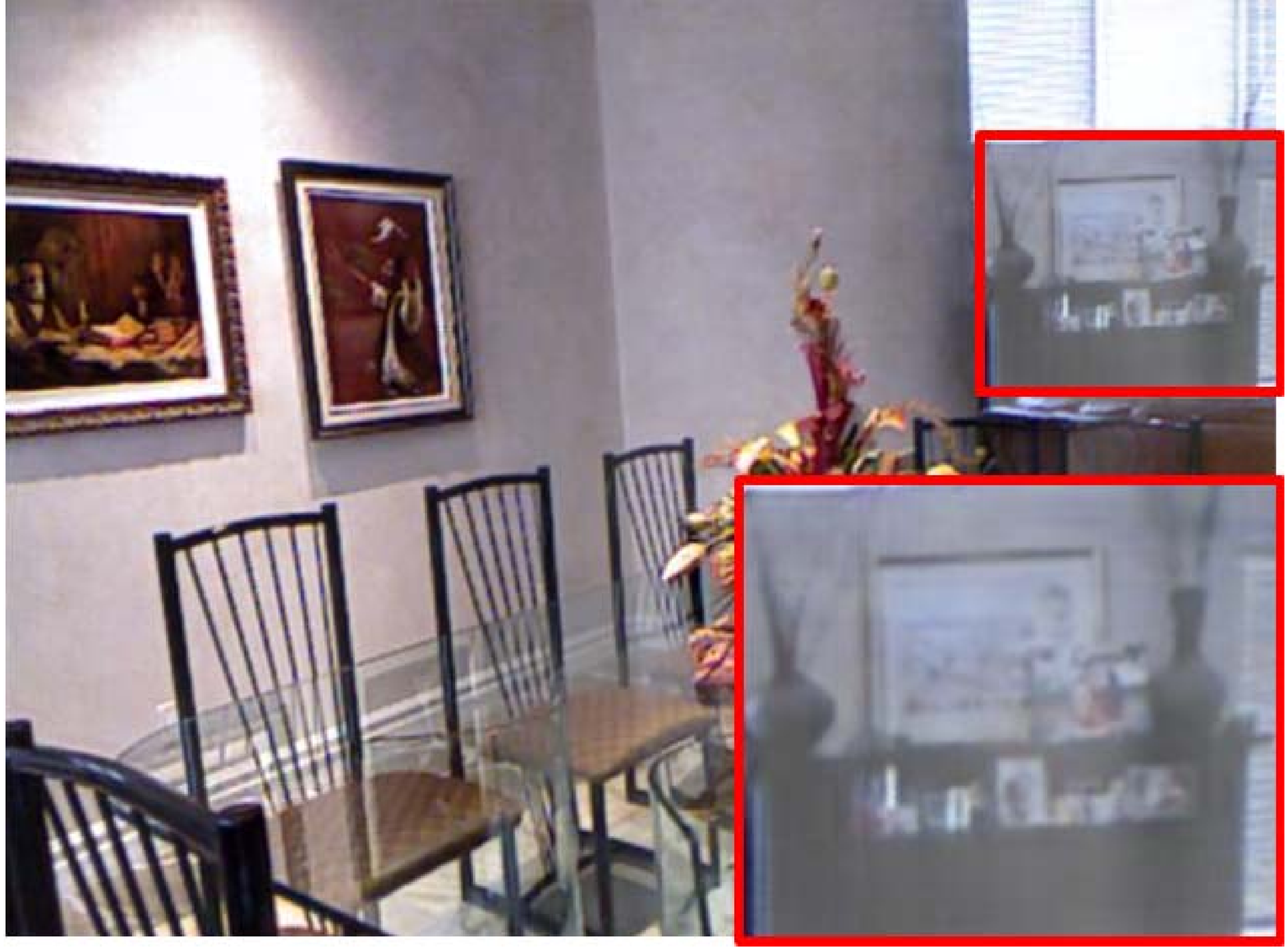} & \hspace{-4mm}
          \includegraphics[width=0.245\linewidth]{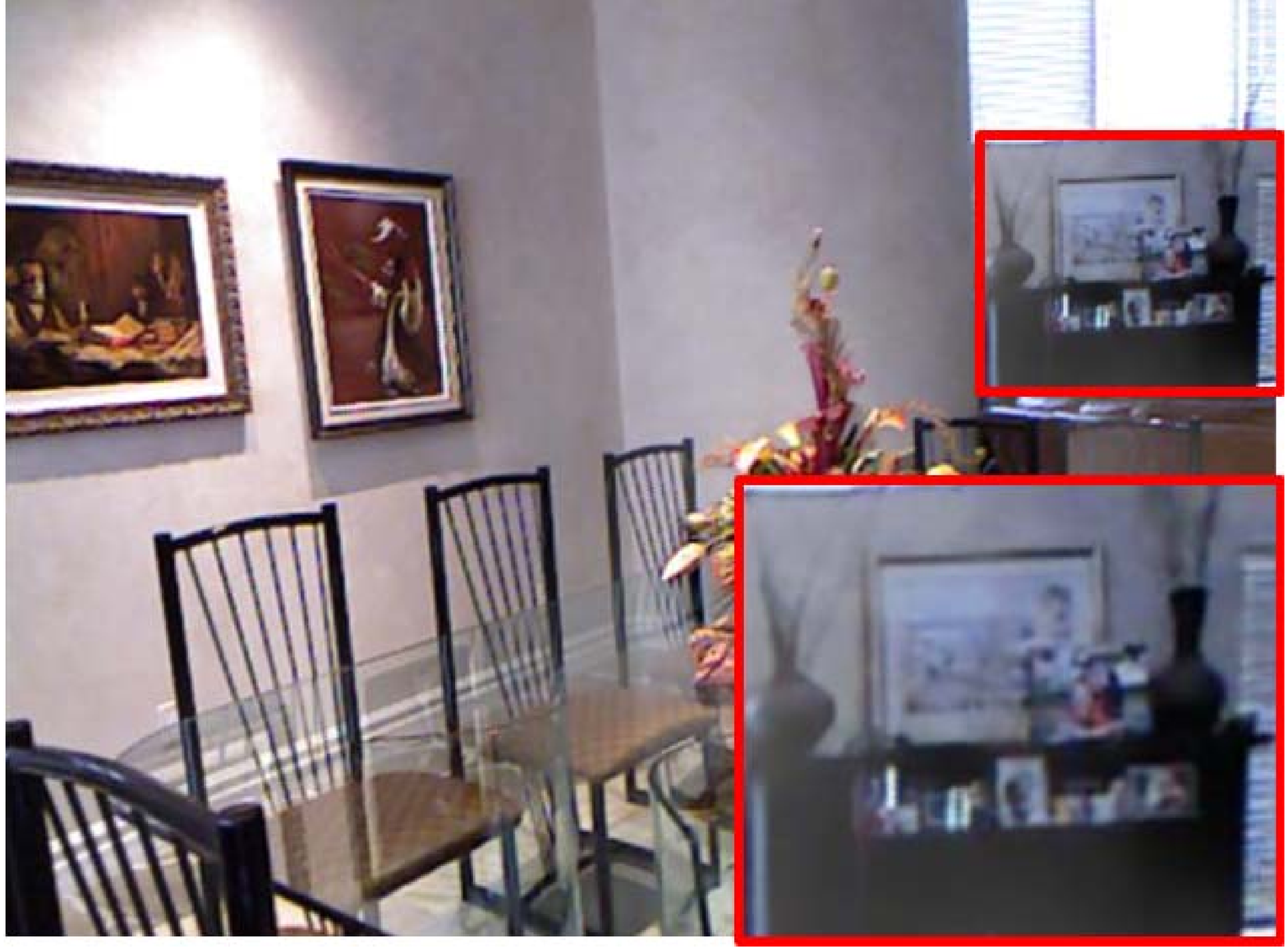} & \hspace{-4mm}
          \includegraphics[width=0.245\linewidth]{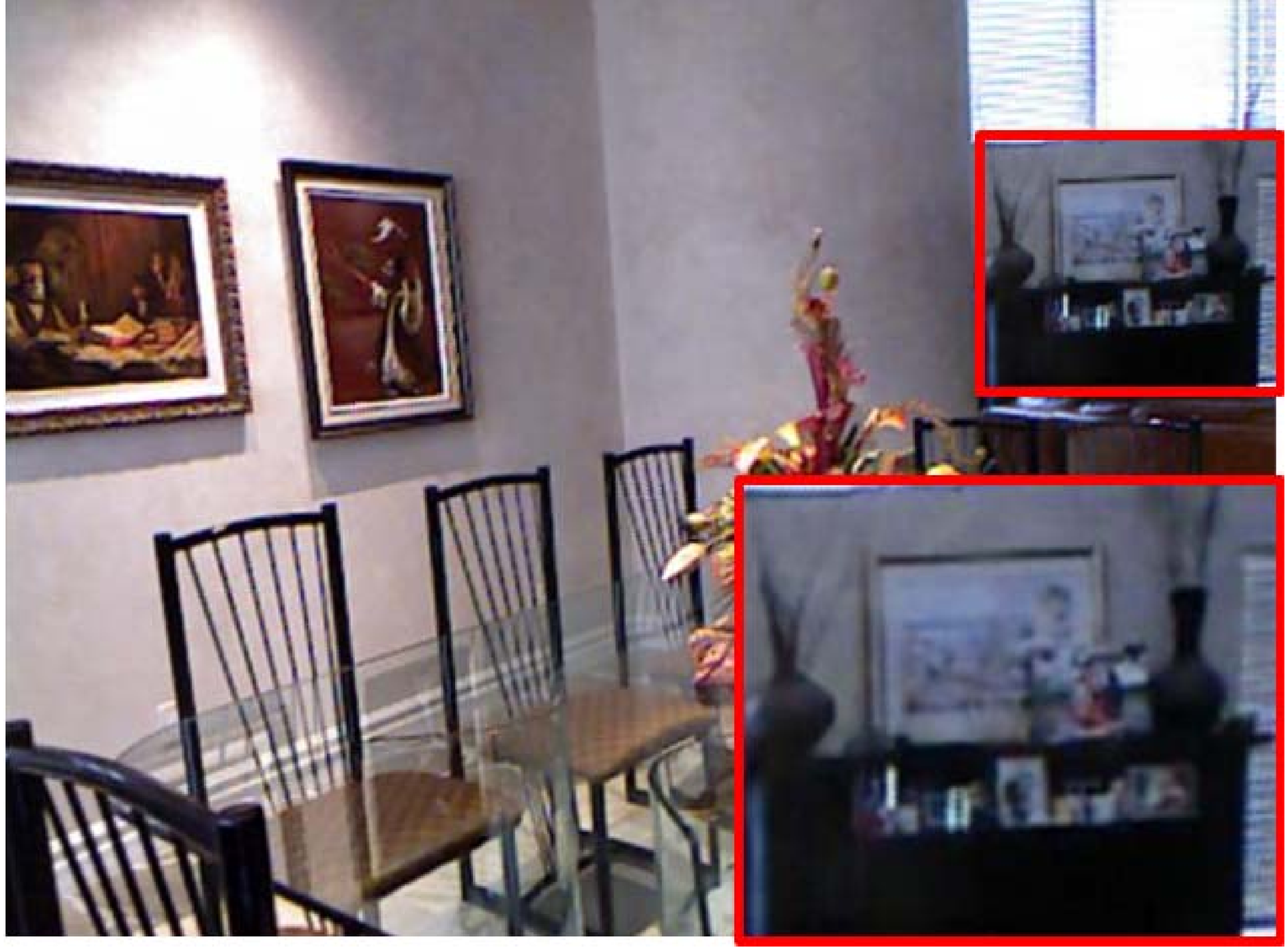}
          \\
          \includegraphics[width=0.245\linewidth]{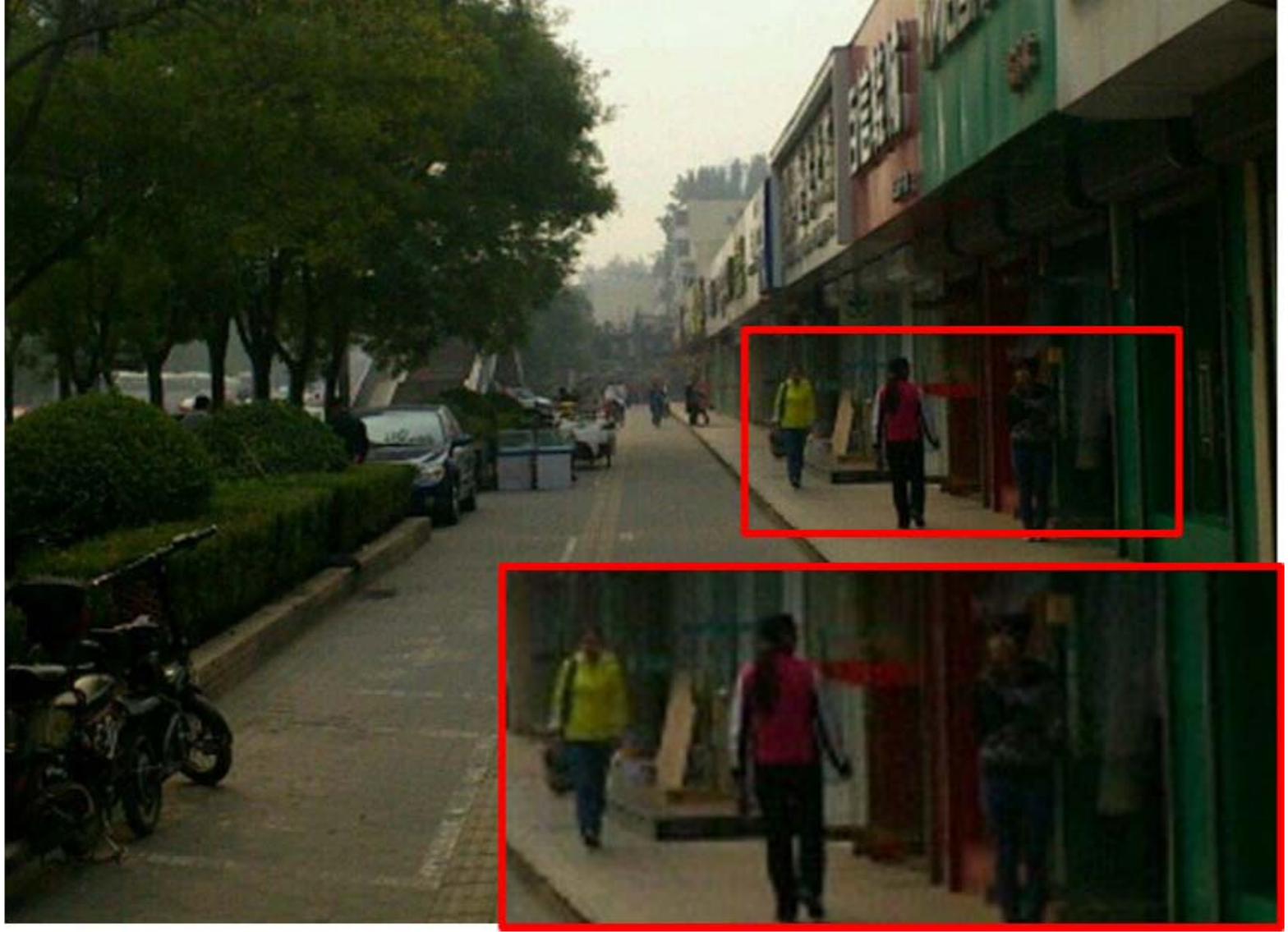} & \hspace{-4mm}
          \includegraphics[width=0.245\linewidth]{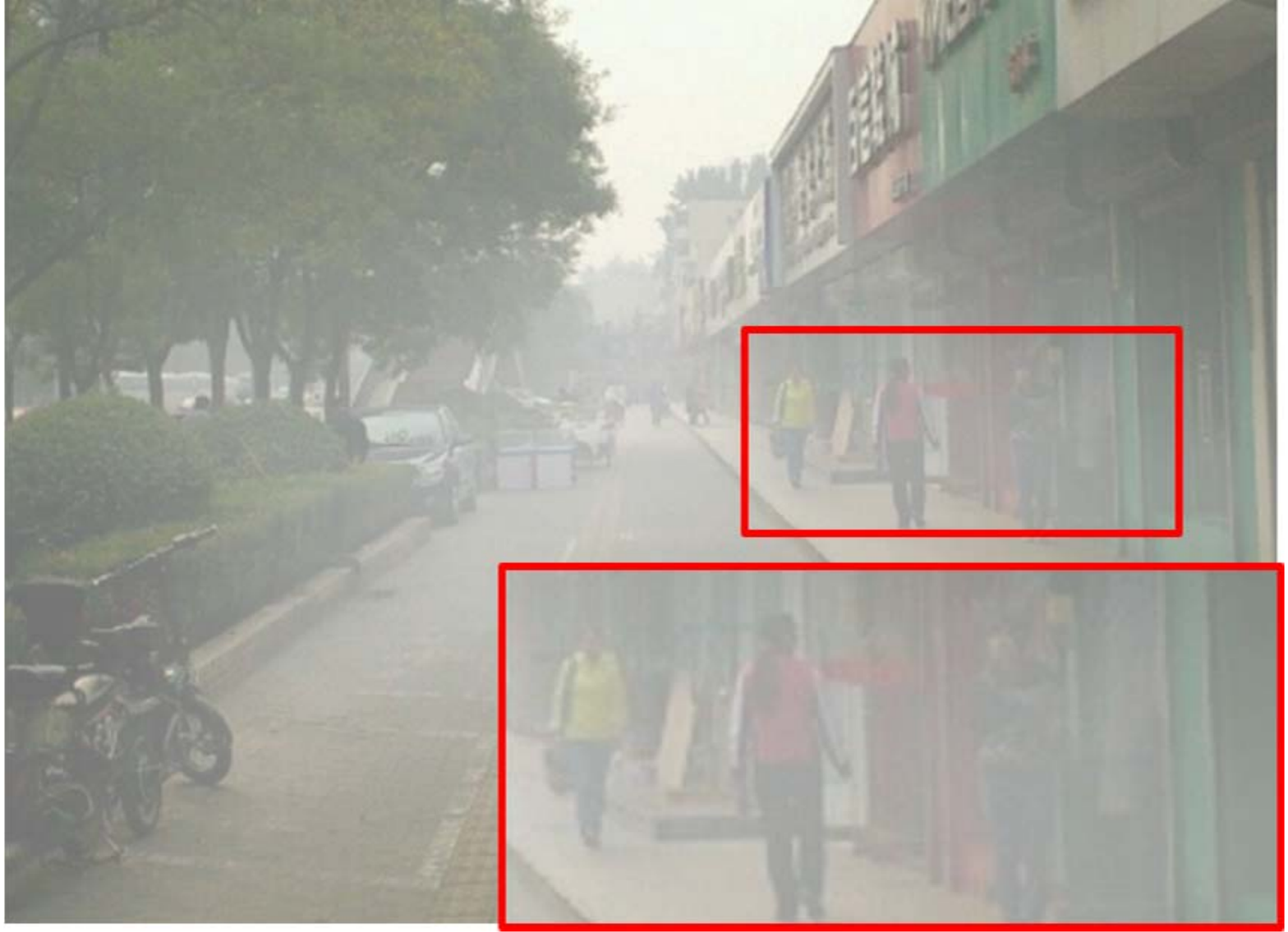} & \hspace{-4mm}
          \includegraphics[width=0.245\linewidth]{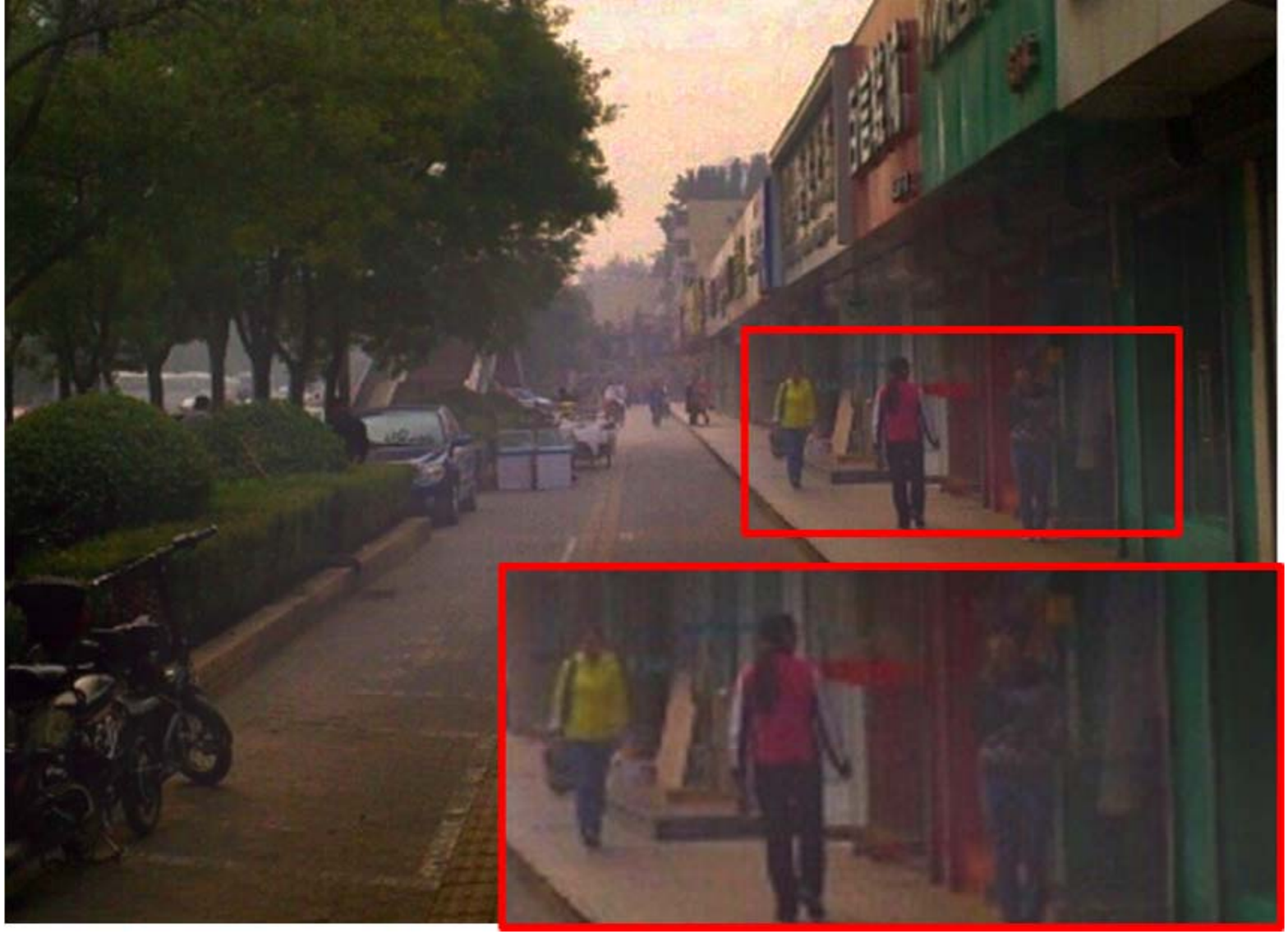} & \hspace{-4mm}
          \includegraphics[width=0.245\linewidth]{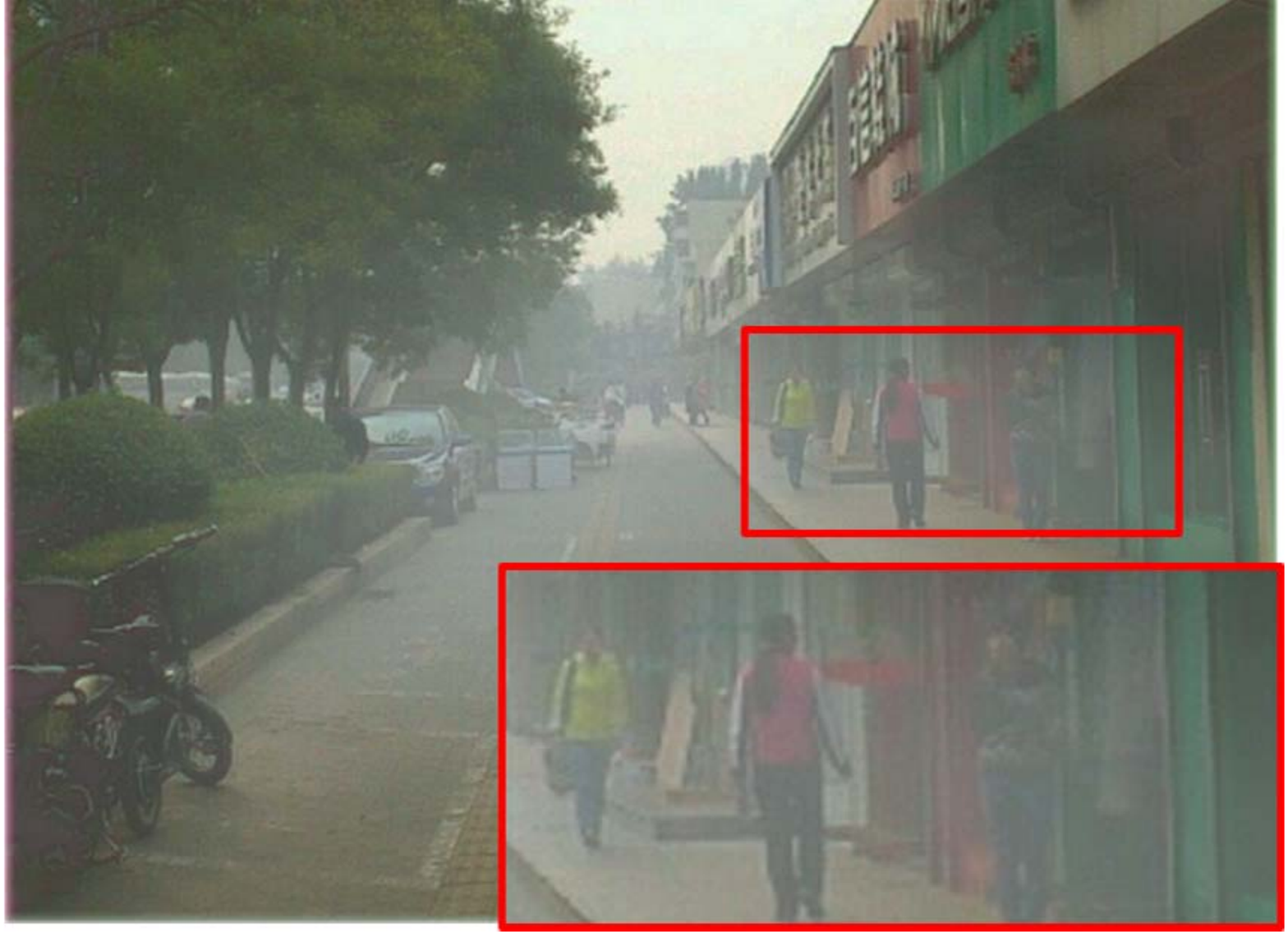} & \hspace{-4mm}
          \includegraphics[width=0.245\linewidth]{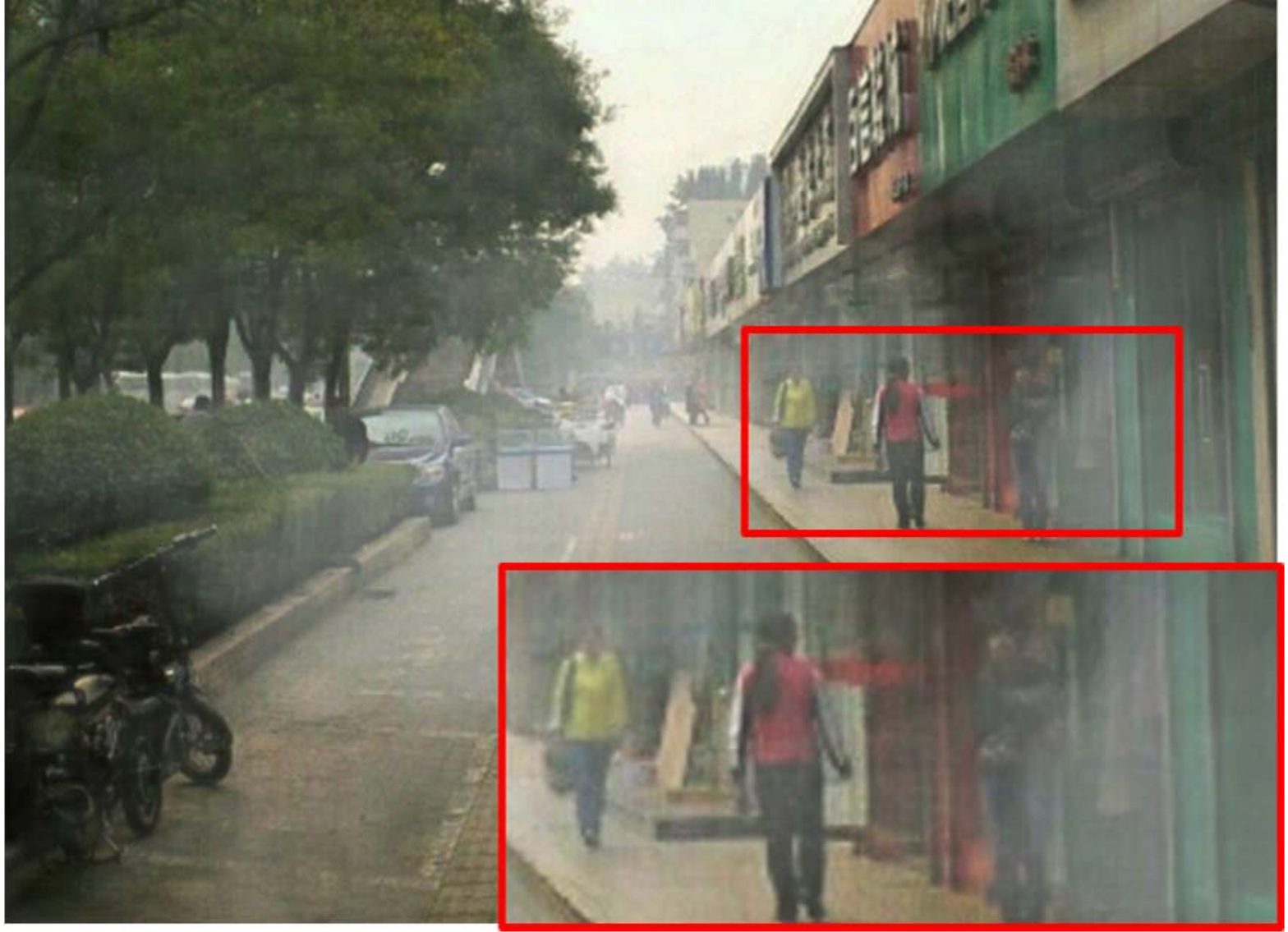} & \hspace{-4mm}
          \includegraphics[width=0.245\linewidth]{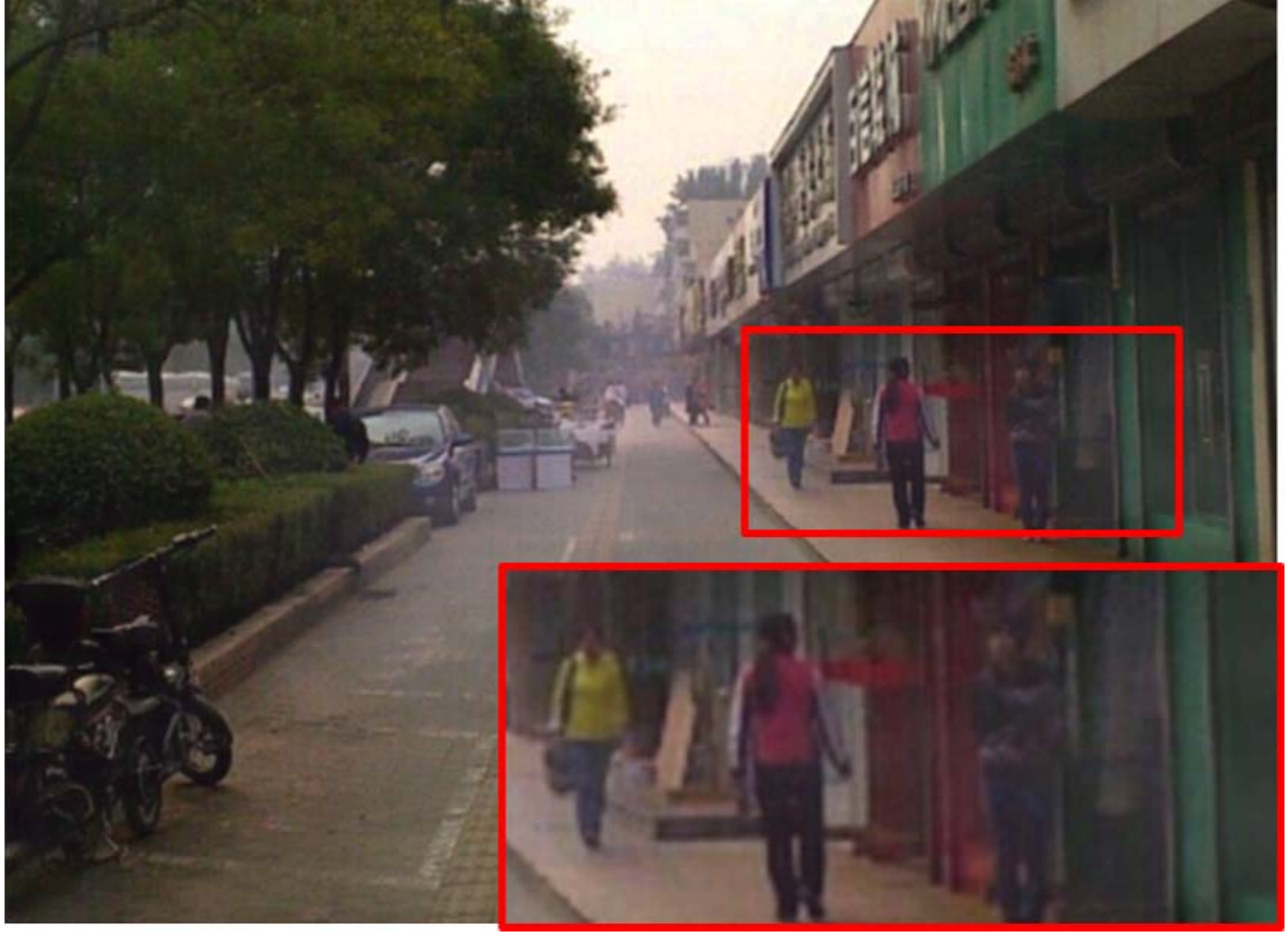} & \hspace{-4mm}
            \includegraphics[width=0.245\linewidth]{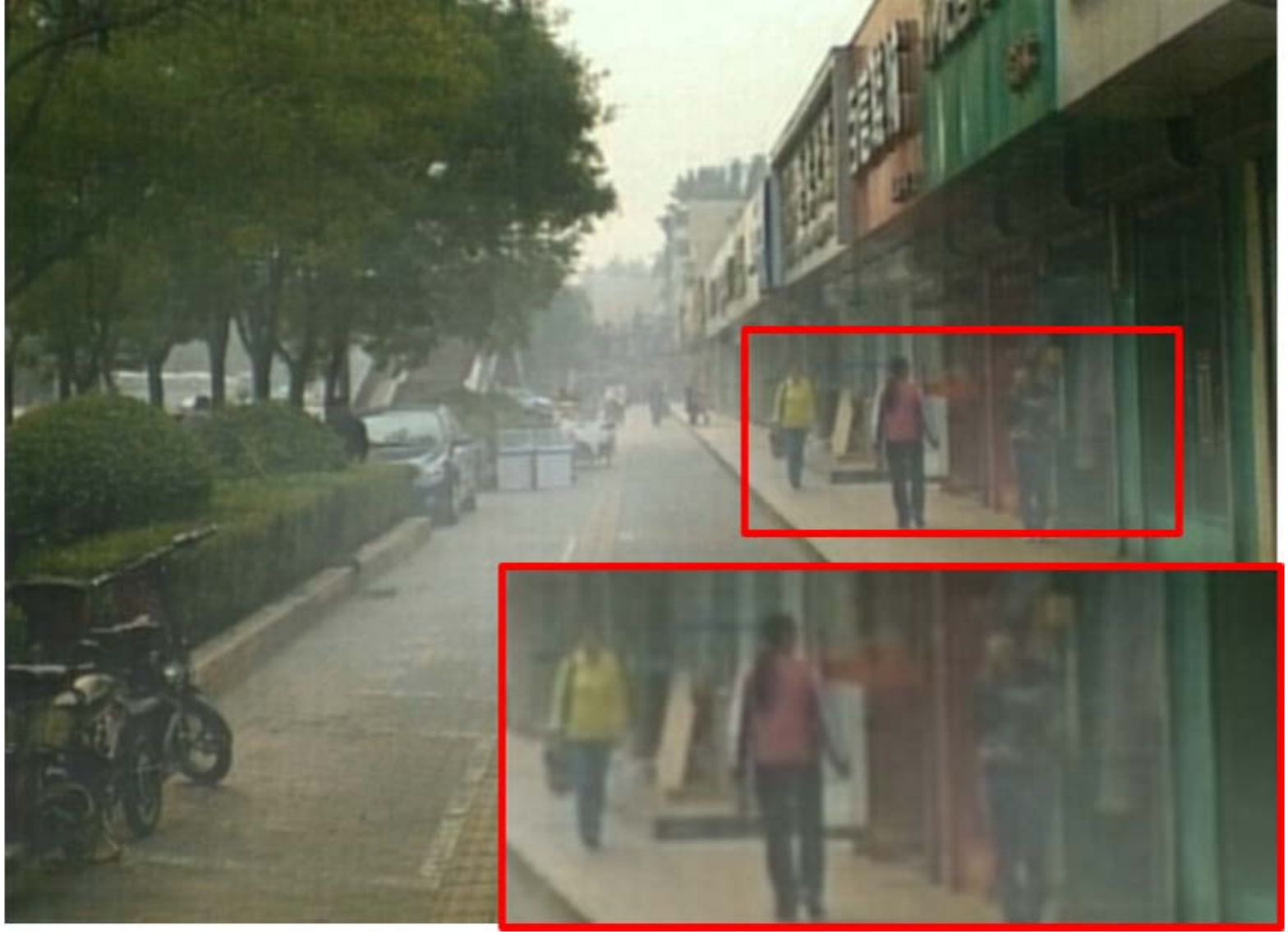} & \hspace{-4mm}
            \includegraphics[width=0.245\linewidth]{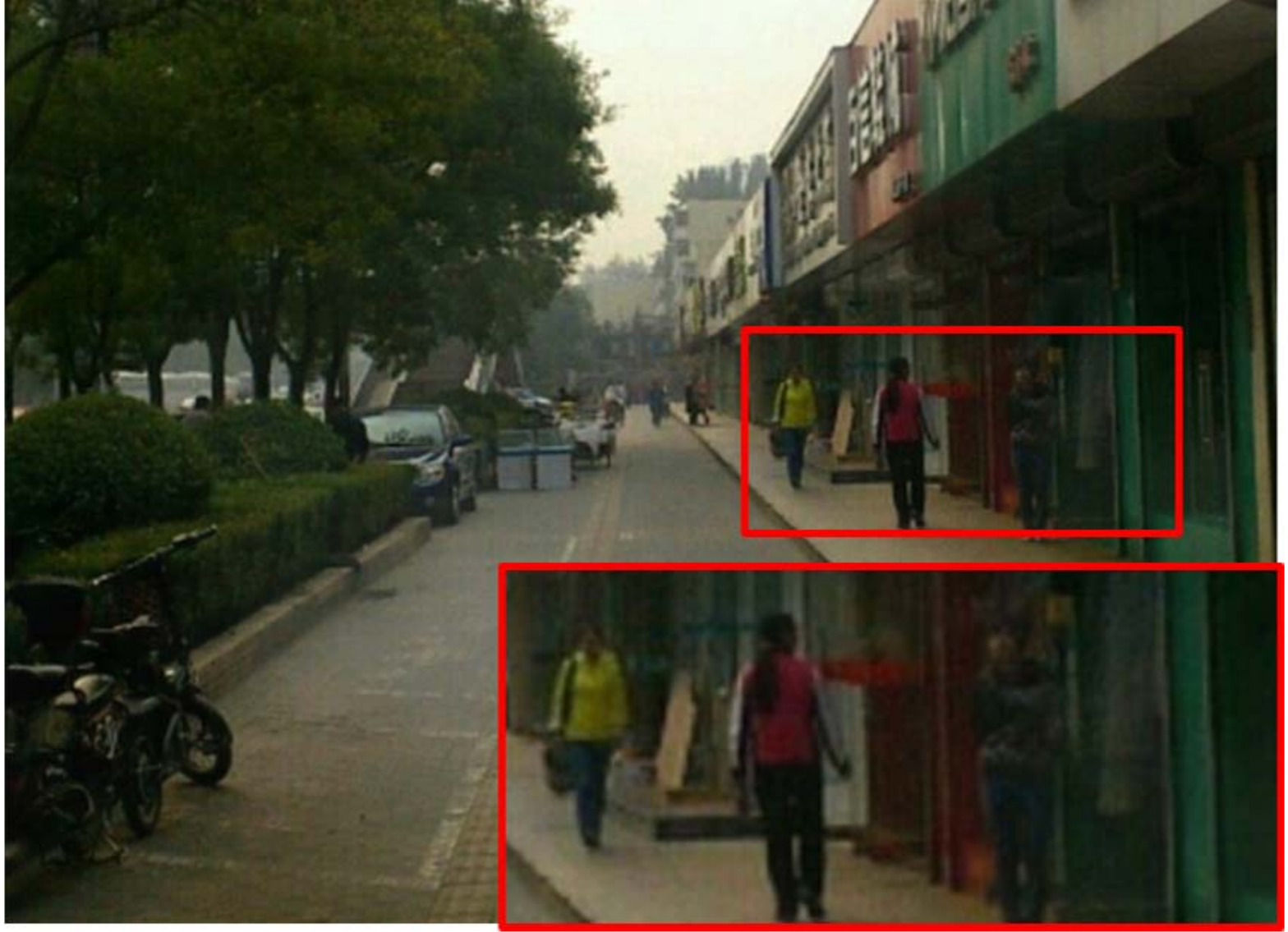}        \\
          \hspace{-3mm}
          (a) Ground-truth & \hspace{-4mm}
          (b) Hazy input & \hspace{-4mm}
          (c) DCP~\cite{He_dark}  & \hspace{-4mm}
          (d) AOD~\cite{AOD} & \hspace{-4mm}
          (e) GFN~\cite{GFN} & \hspace{-4mm}
          (f) GCANet~\cite{GCANet} & \hspace{-4mm}
          (g) DuRN~\cite{DuRN}  & \hspace{-4mm}
          (h) Ours \\
      \end{tabular}
    \end{adjustbox}
    \caption{ \textbf{Visual results on the SOTS dataset.}
    The results in (c)-(g) contain some color distortions and haze residual, while the dehazed image in (h) by our method is much clearer.
    Best viewed on a high-resolution display.}
    \label{fig:visual_results_SOTS}
    \vspace{-3mm}
    \end{figure*}

  \figref{visual_results_SOTS} shows two examples from the SOTS dataset. The DCP algorithm generates the results with significant color distortions.
  The dehazed images by other deep learning frameworks still contain significant artifacts.
  In contrast, our algorithm restores these images well.
  
\begin{figure*}[!t]
    \large
    \centering
    \begin{adjustbox}{width=\linewidth}
      \begin{tabular}{cccccc}
      \includegraphics[width=0.245\linewidth]{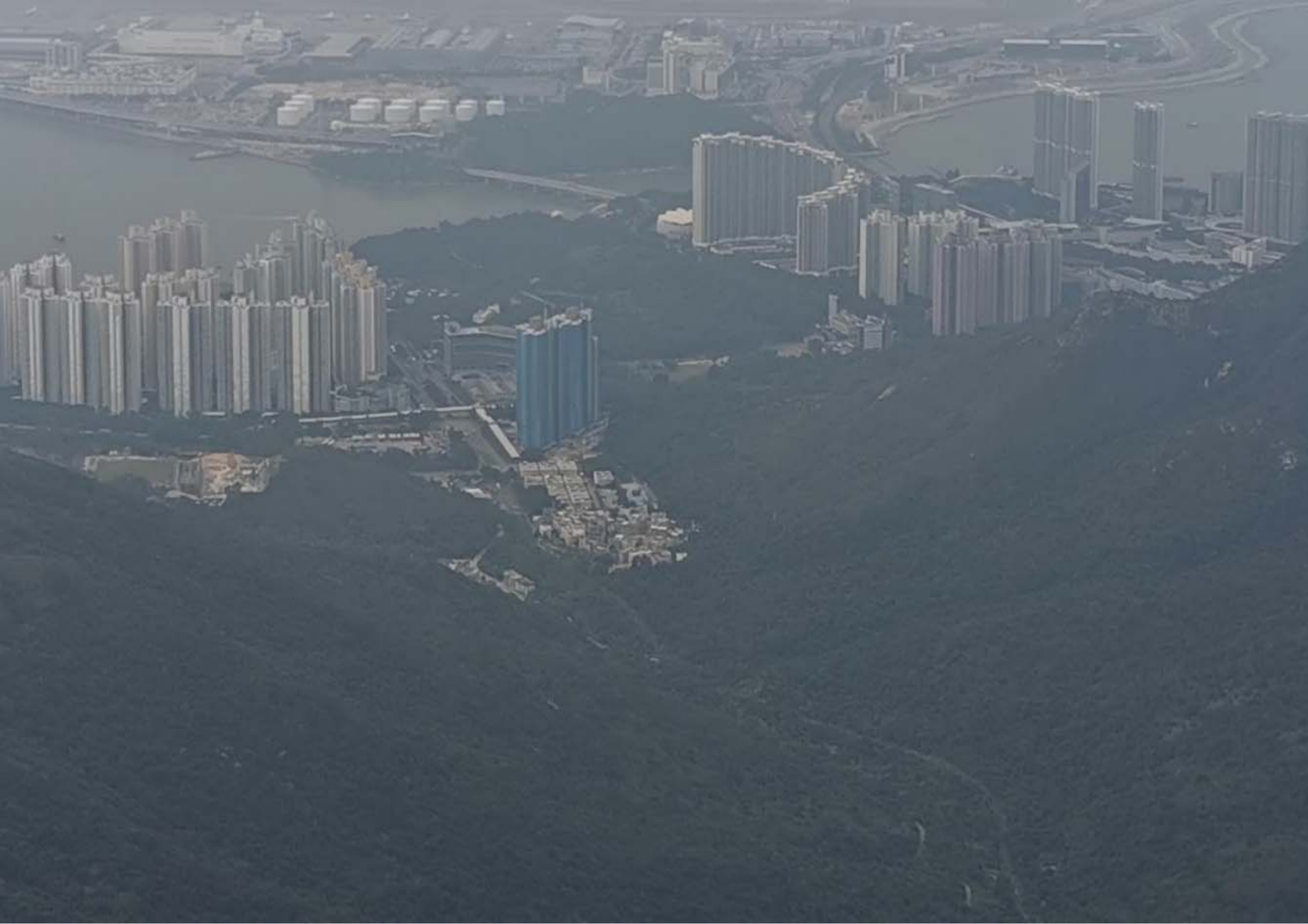} & \hspace{-4mm}
      \includegraphics[width=0.245\linewidth]{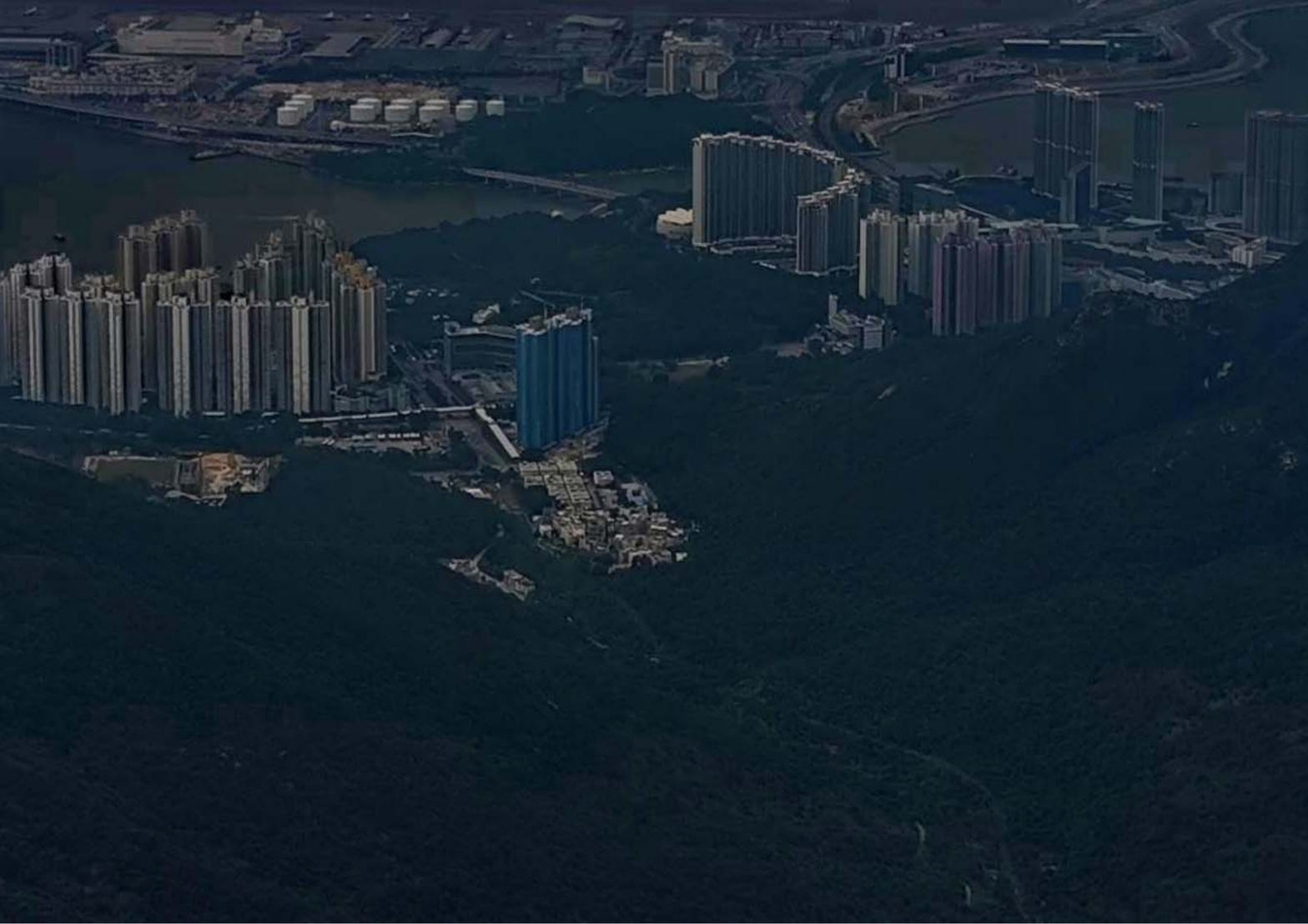} & \hspace{-4mm}
      \includegraphics[width=0.245\linewidth]{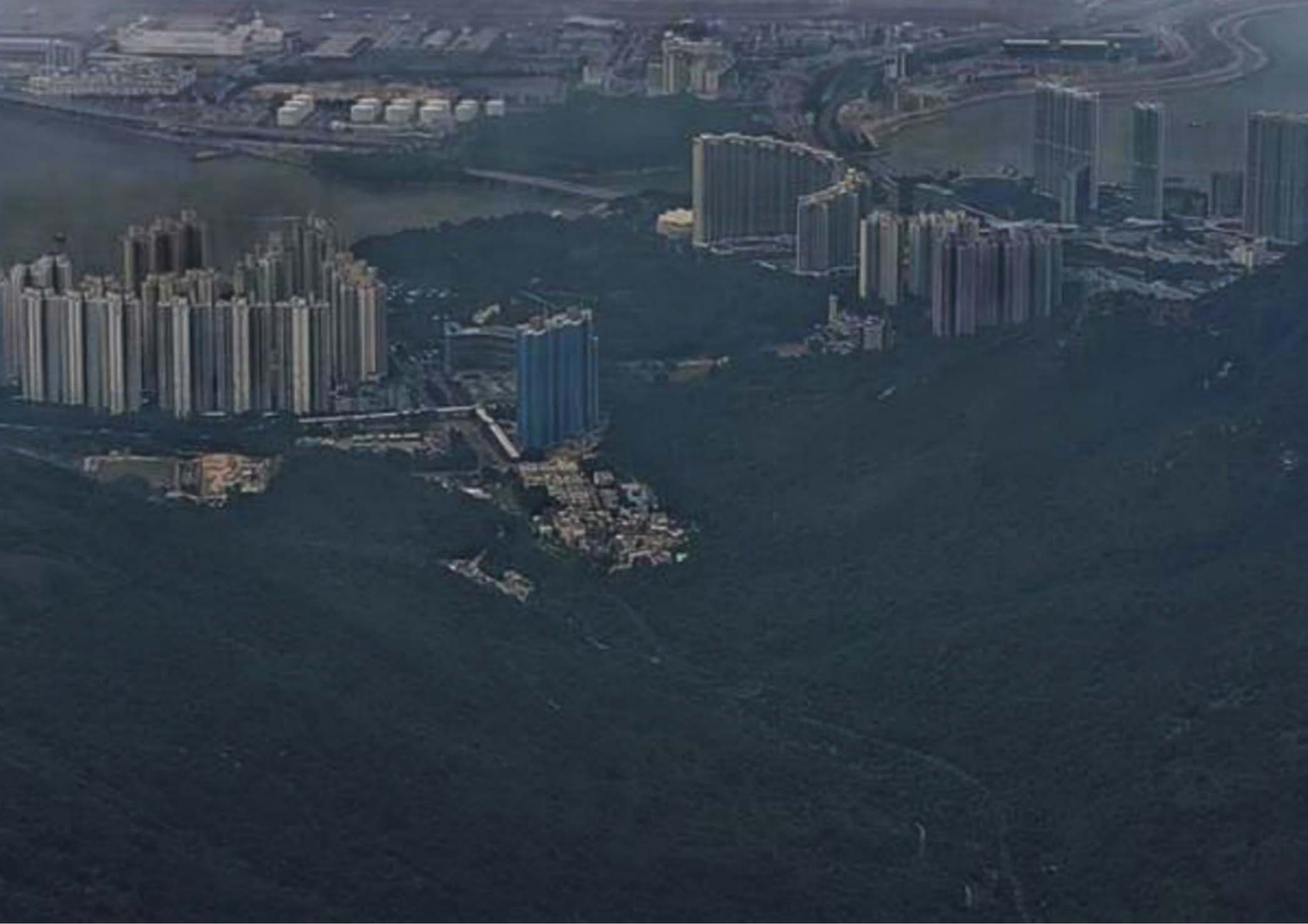} & \hspace{-4mm}
      \includegraphics[width=0.245\linewidth]{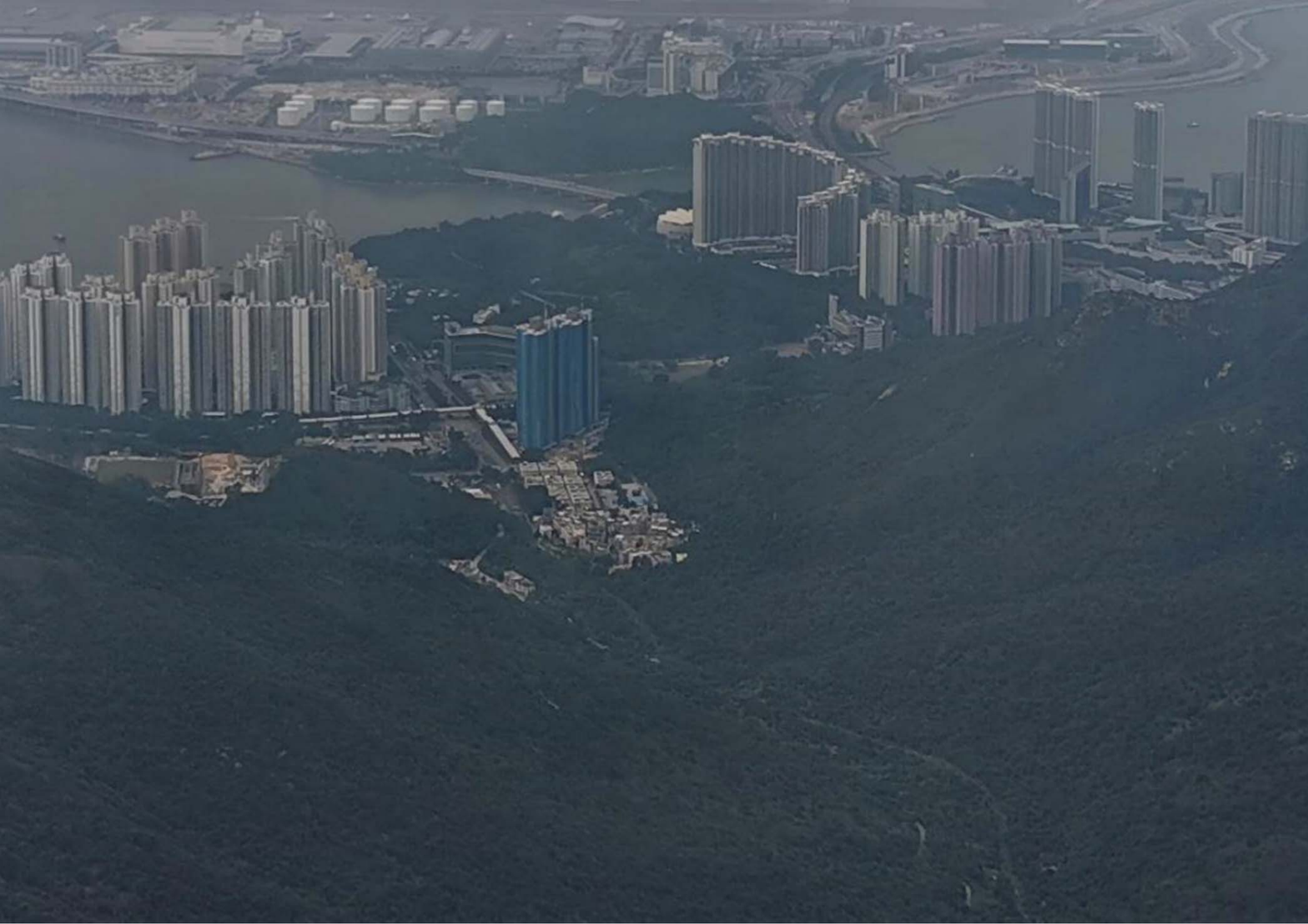} & \hspace{-4mm}
        \includegraphics[width=0.245\linewidth]{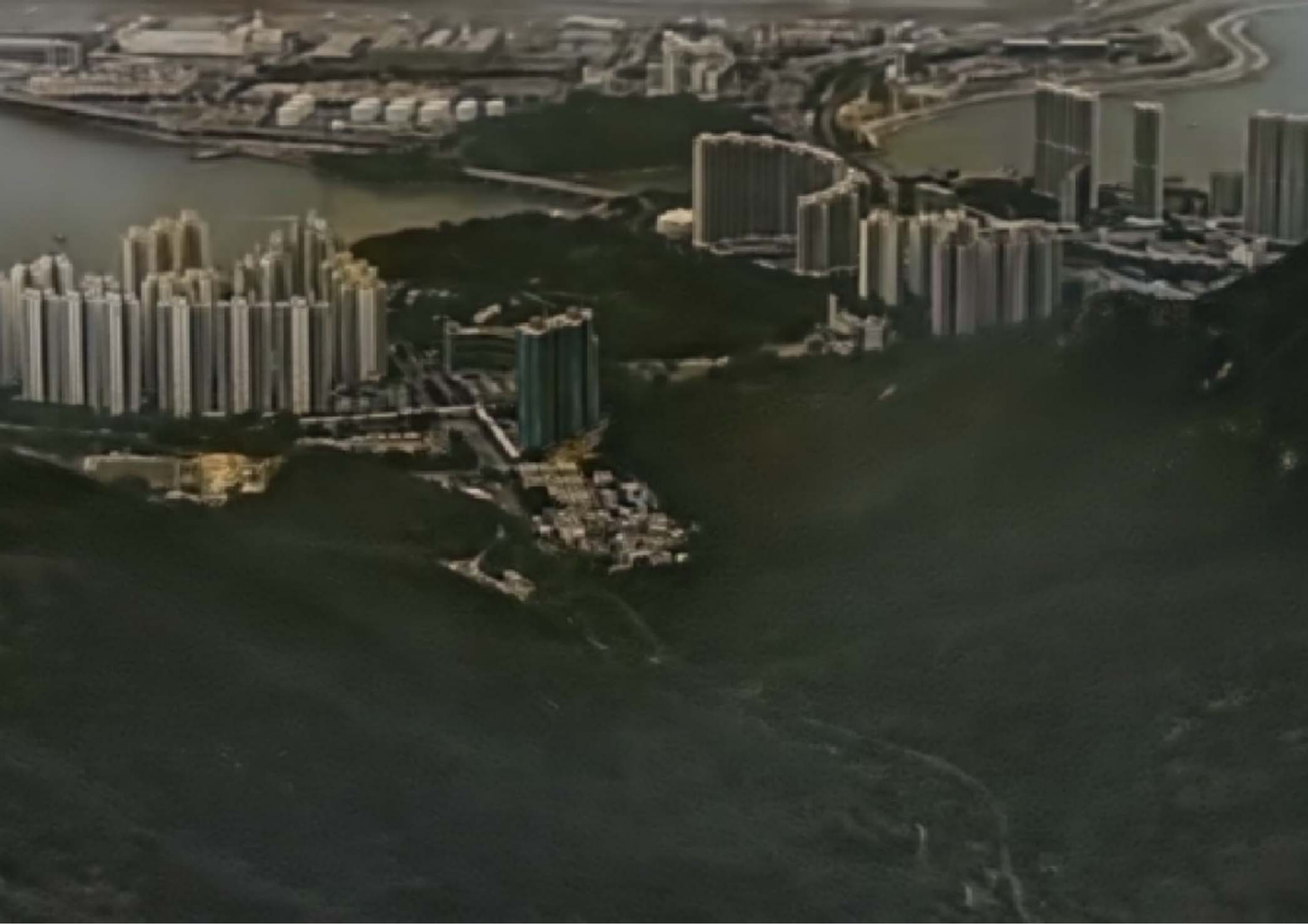} & \hspace{-4mm}
        \includegraphics[width=0.245\linewidth]{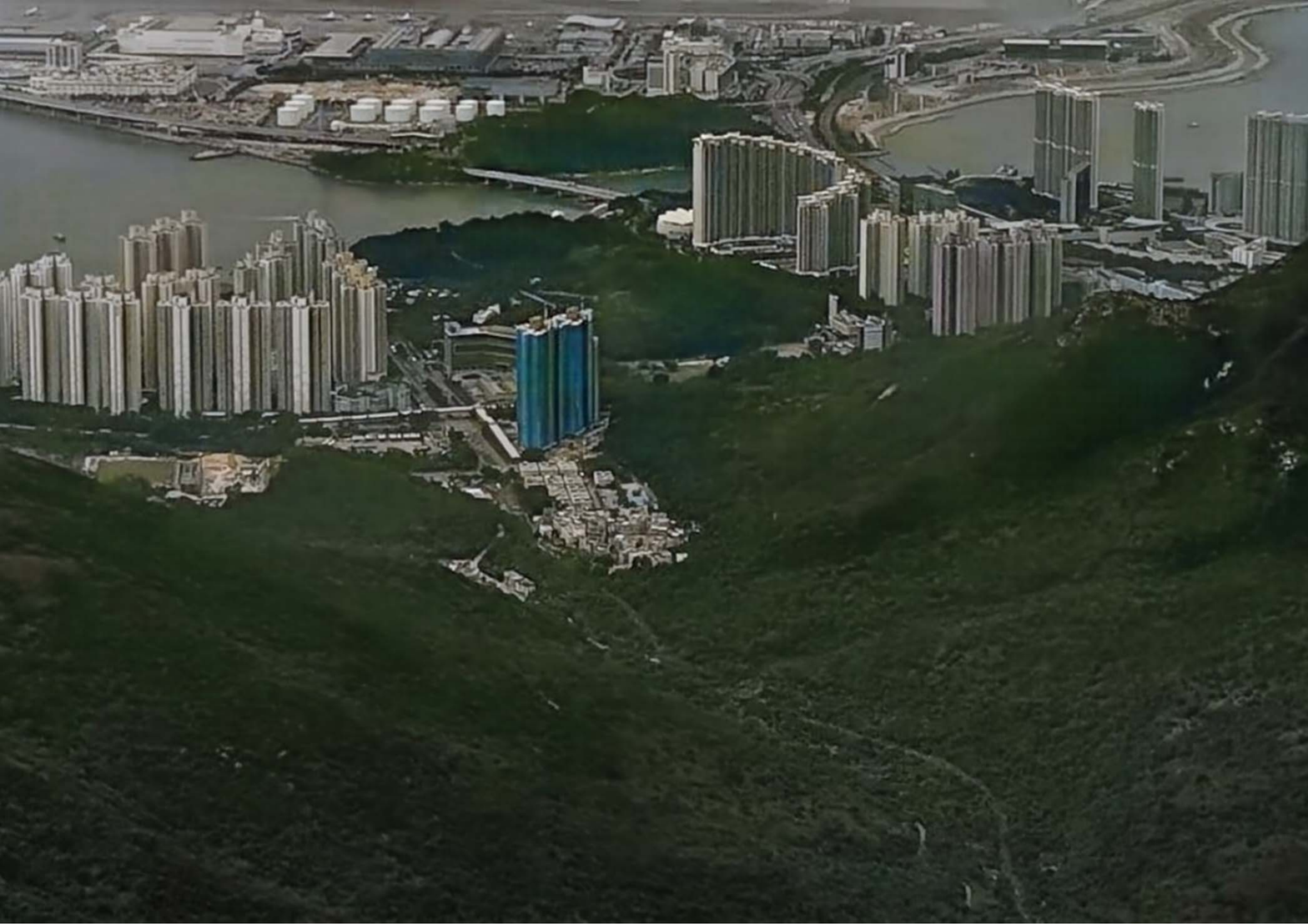}\\
  \hspace{-3mm}
          (a) Hazy input & \hspace{-4mm}
          (b) DCP~\cite{He_dark}  & \hspace{-4mm}
          (c) GFN~\cite{GFN} & \hspace{-4mm}
          (d) PFFNet~\cite{PFFNet} & \hspace{-4mm}
          (e) DuRN~\cite{DuRN}  & \hspace{-4mm}
          (f) Ours \\
      \end{tabular}
    \end{adjustbox}
    \caption{
    \textbf{Visual results on the real-world image.}
    The proposed method generates a clearer dehazed image with less color distortions.
    Best viewed on a high-resolution display.
    }
    \label{fig:visual_results_real}
    \vspace{-3mm}
    \end{figure*}

  We further evaluate our algorithm on real images.
  \figref{visual_results_real} shows a real hazy image and the dehazed results from state-of-the-art methods.
  The dehazed image by our method is sharper and brighter.
  More qualitative results are provided in the supplementary material.
  
  %
  \noindent{\bf Perceptual quality for high-level vision tasks.}
  As the dehazing algorithms are usually used as the pre-processing step for high-level applications, it is helpful to evaluate the perceptual quality of the dehazed results.
  In this work, we provide evaluation results on the perceptual quality for the object detection task.
  To obtain the data for this evaluation, we generate hazy images using the images from the KITTI detection dataset~\cite{kitti_detection}.
  We first estimate the depth map for each image by a single-image depth estimation method Monodepth2~\cite{monodepth2}, and then use the depth map to synthesize a haze image following the protocols of the RESIDE dataset~\cite{RESIDE}.
  This dataset is referred to as the KITTI Haze dataset in this work.
  We evaluate the proposed method with the following approaches: DCP~\cite{He_dark}, GFN~\cite{GFN}, PFFNet~\cite{PFFNet}, and DuRN~\cite{DuRN}.
  For detection accuracy, we use the state-of-the-art method YOLOv3~\cite{yolov3} to evaluate on the dehazed images from different dehazing methods.
  
\begin{figure*}[!t]
  \footnotesize
  \centering
  \begin{tabular}{ccccc}
        \includegraphics[width=0.19\linewidth]{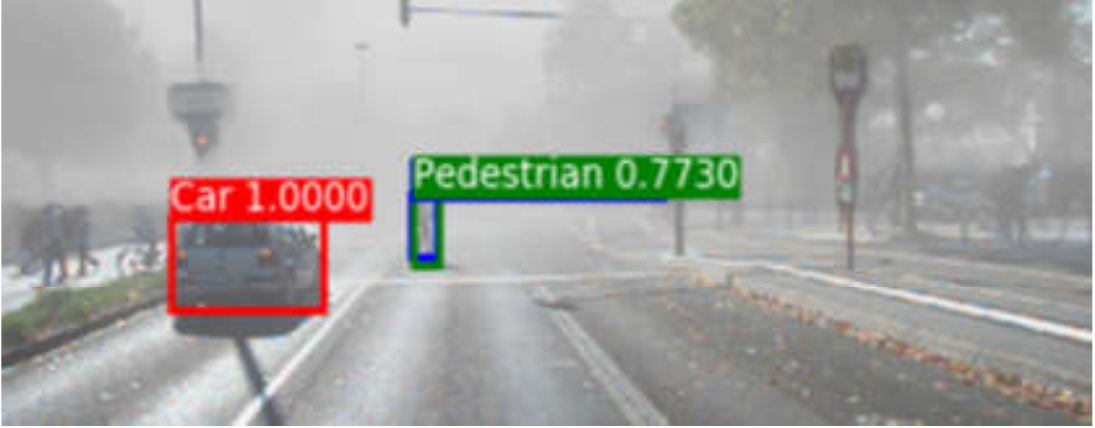} & \hspace{-4mm}
        \includegraphics[width=0.19\linewidth]{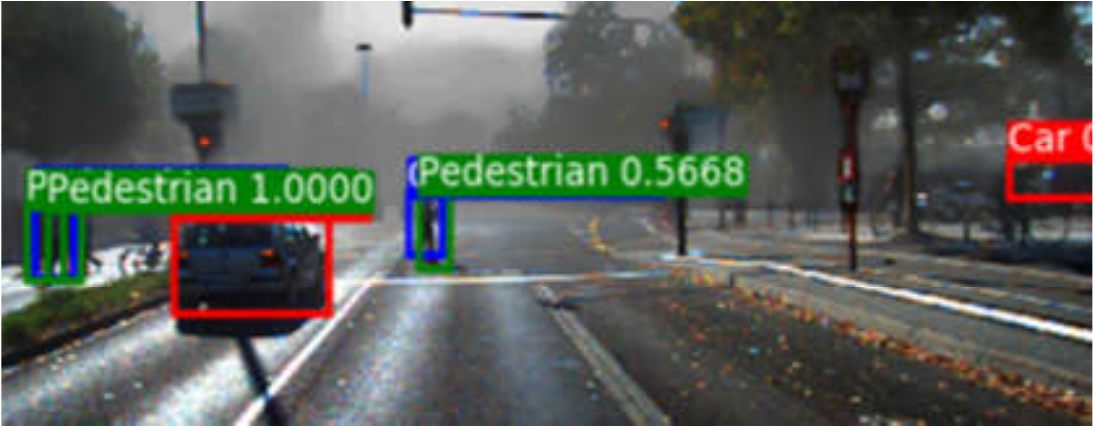} & \hspace{-4mm}
        \includegraphics[width=0.19\linewidth]{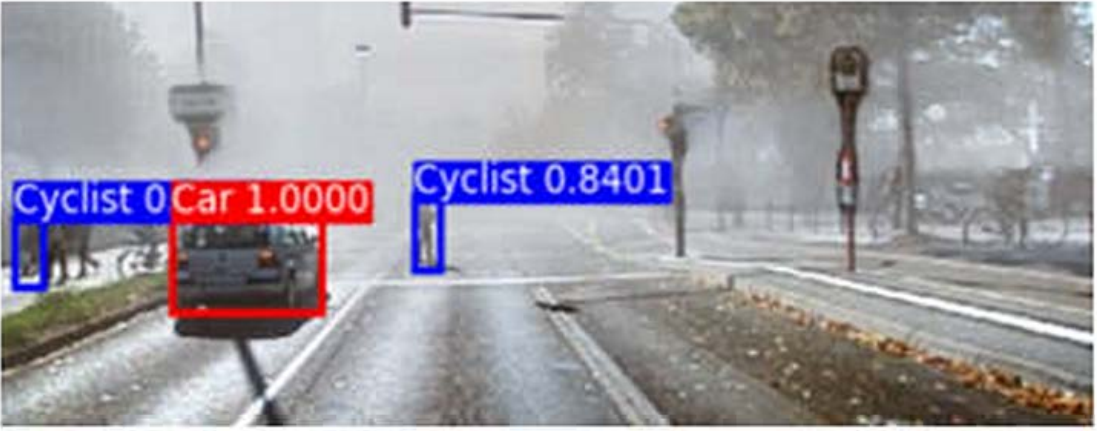} & \hspace{-4mm}
        \includegraphics[width=0.19\linewidth]{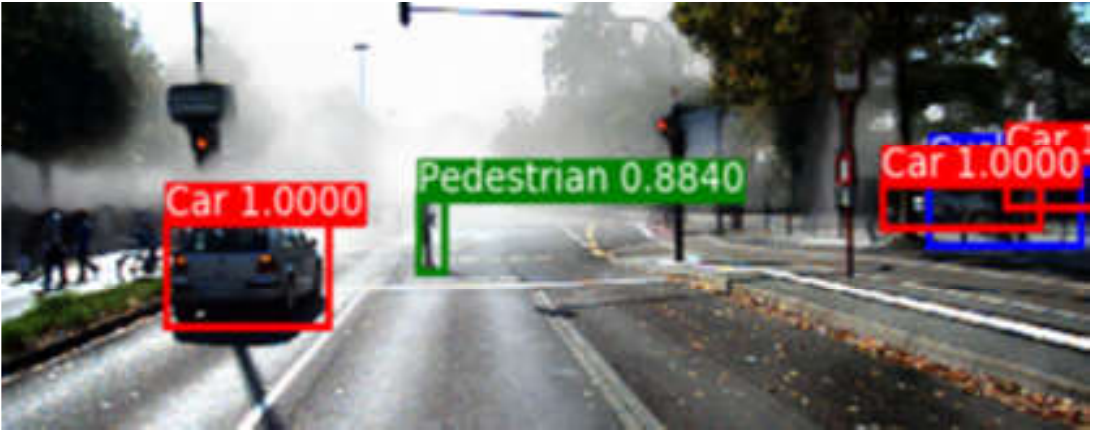} & \hspace{-4mm}
        \includegraphics[width=0.19\linewidth]{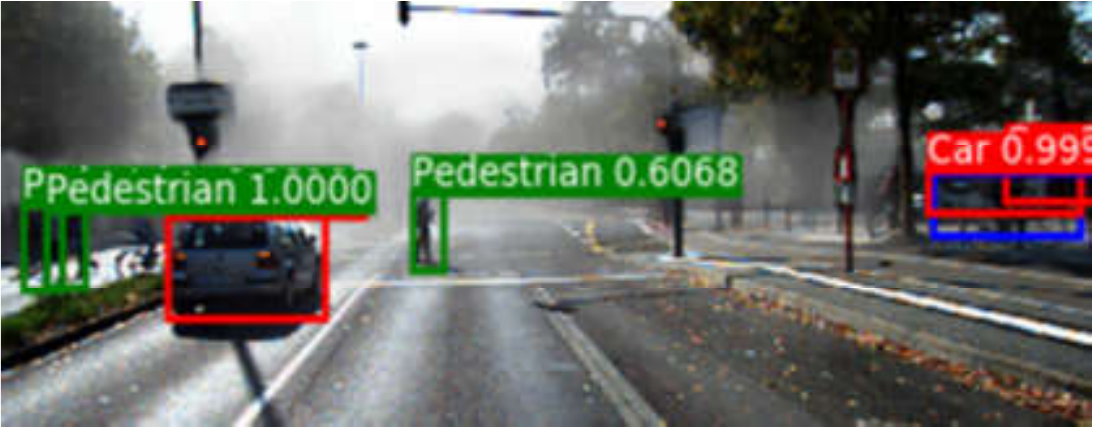}
     \\
        (a) Hazy & \hspace{-4mm}
        (b) DCP~\cite{He_dark} & \hspace{-4mm}
        (c) PFFNet~\cite{PFFNet} & \hspace{-4mm}
        (d) DuRN~\cite{DuRN} & \hspace{-4mm}
        (e) Ours
  \end{tabular}
  \caption{
  \textbf{Detection results using the dehazed images from the synthesized KITTI Haze dataset.}
  Best viewed on a high-resolution display.
  }
  \label{fig:visual_results_detection_hazy}
  \vspace{-3mm}
  \end{figure*}

  The detection results are shown in \tabref{heavy_kitti}.
  The dehazed images restored from the proposed method obtain the highest detection accuracy, which shows that our method can restore images with better perceptual quality.
  The qualitative results in \figref{visual_results_detection_hazy} demonstrate
  that our method not only generates better dehazed image but also helps the detection algorithm to recognize cars and pedestrians.
  
  \subsection{Ablation Study and Analysis}

  \label{sec:4.2}
  In this section, we analyze how the proposed method performs for image dehazing.
  All the baseline methods mentioned below are trained using the same setting as the proposed algorithm for fair comparisons.
  %
  \begin{figure*}[!t]
     \centering
     \LARGE
     \begin{adjustbox}{width=0.95\linewidth}
       \begin{tabular}{ccccc}
       \hspace{-3mm}
       \includegraphics[width=0.4\linewidth]{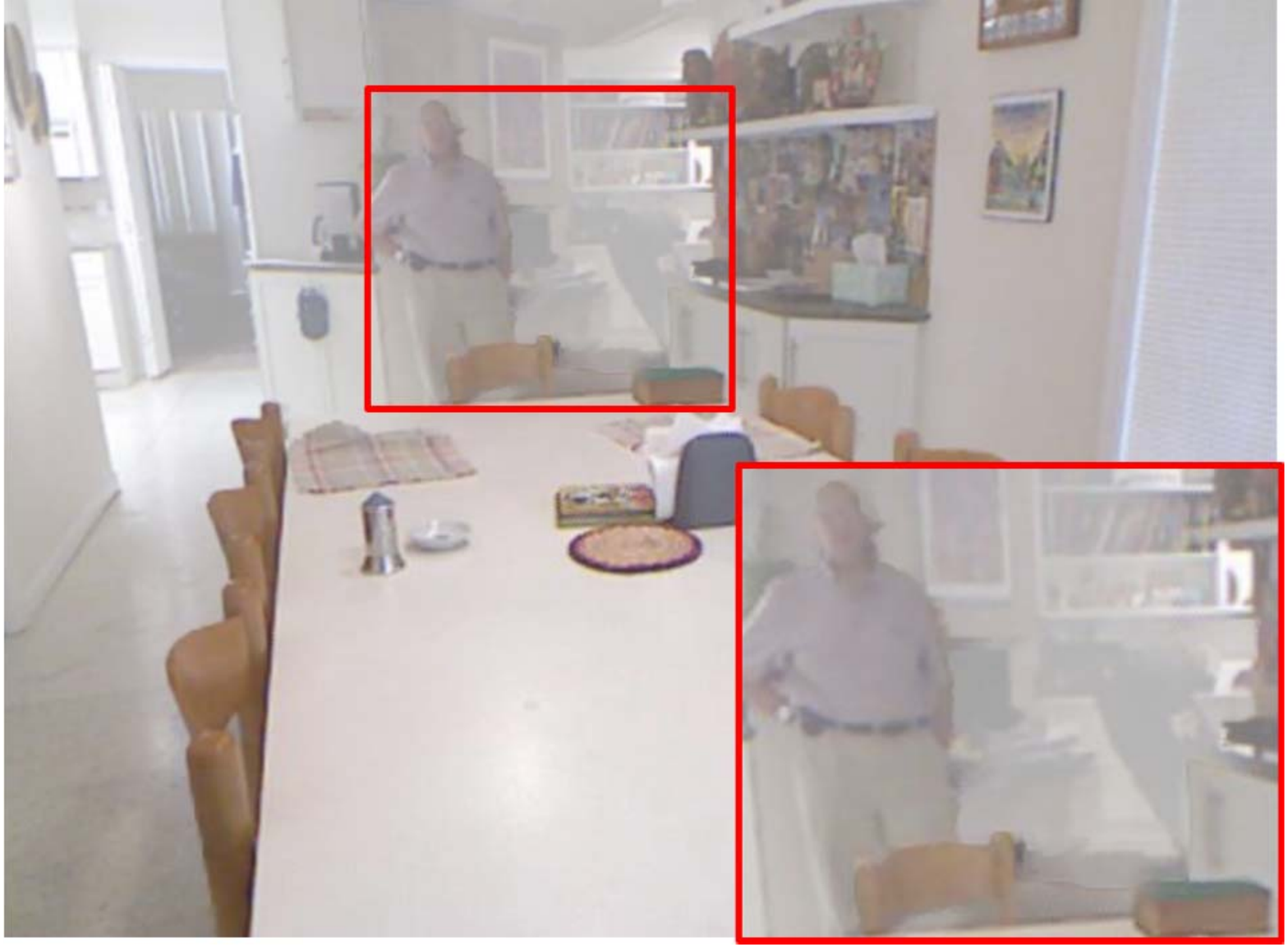} & \hspace{-4mm}
       \includegraphics[width=0.4\linewidth]{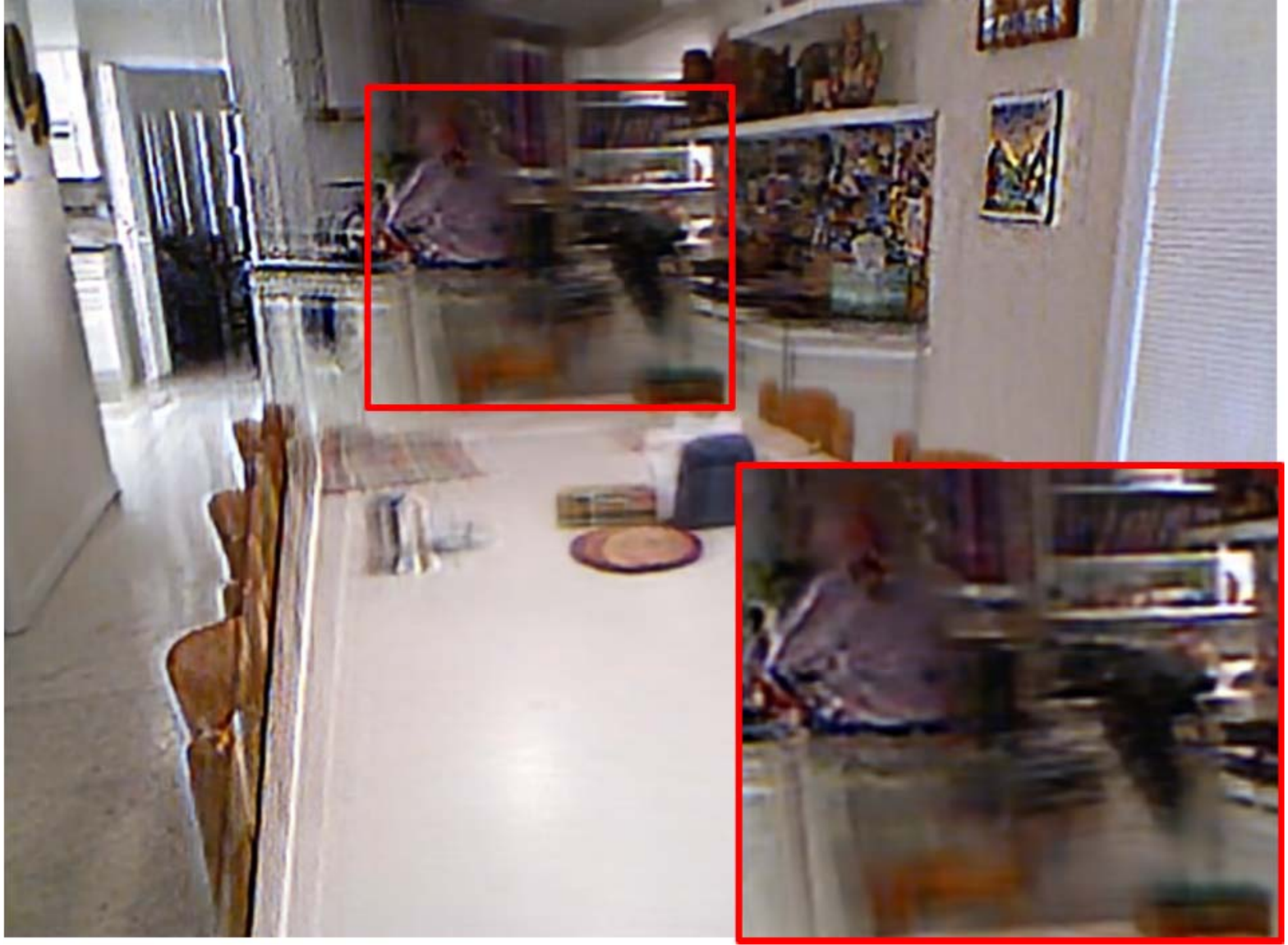} & \hspace{-4mm}
       \includegraphics[width=0.4\linewidth]{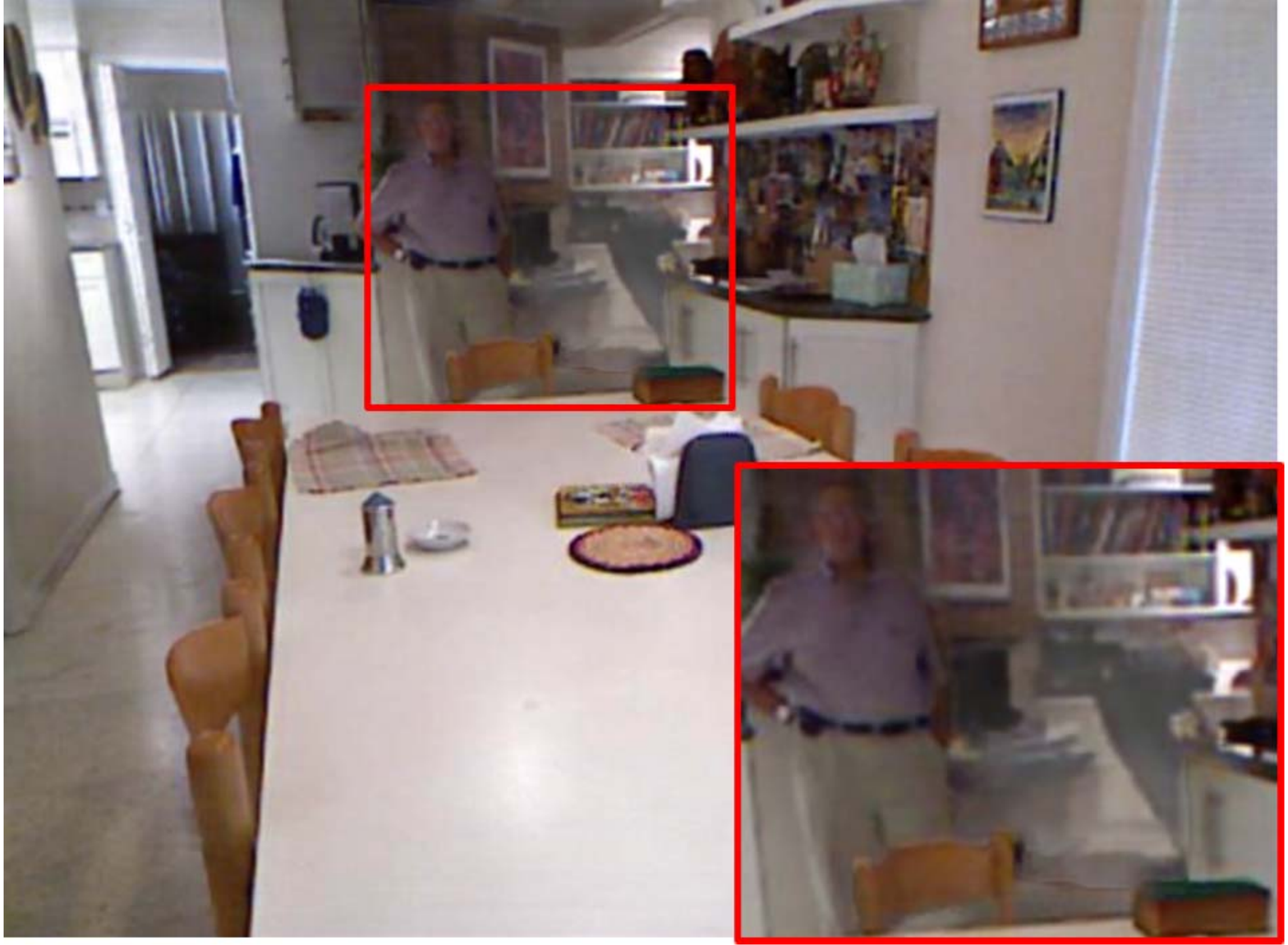} & \hspace{-4mm}
       \includegraphics[width=0.4\linewidth]{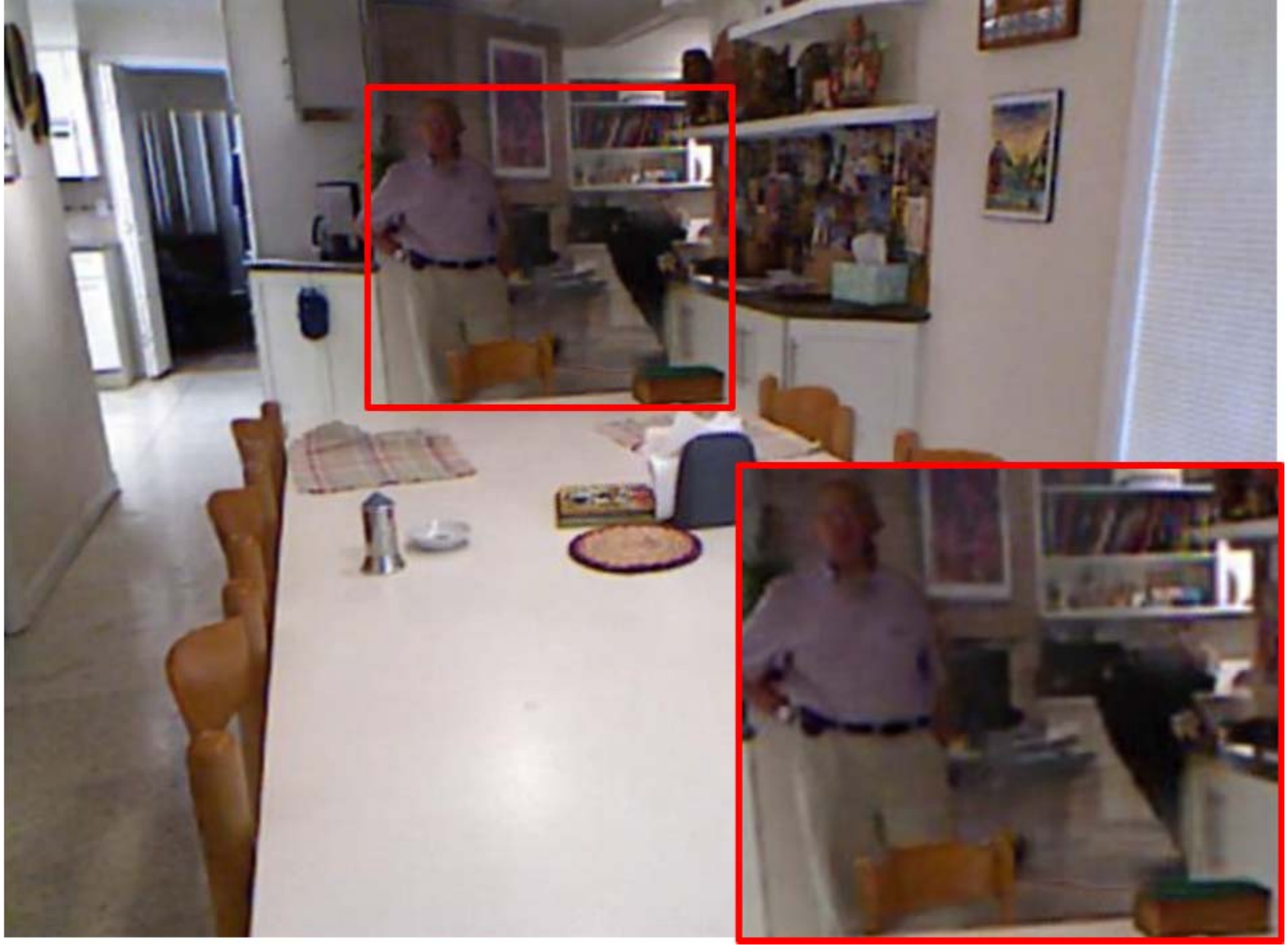} & \hspace{-4mm}
         \includegraphics[width=0.4\linewidth]{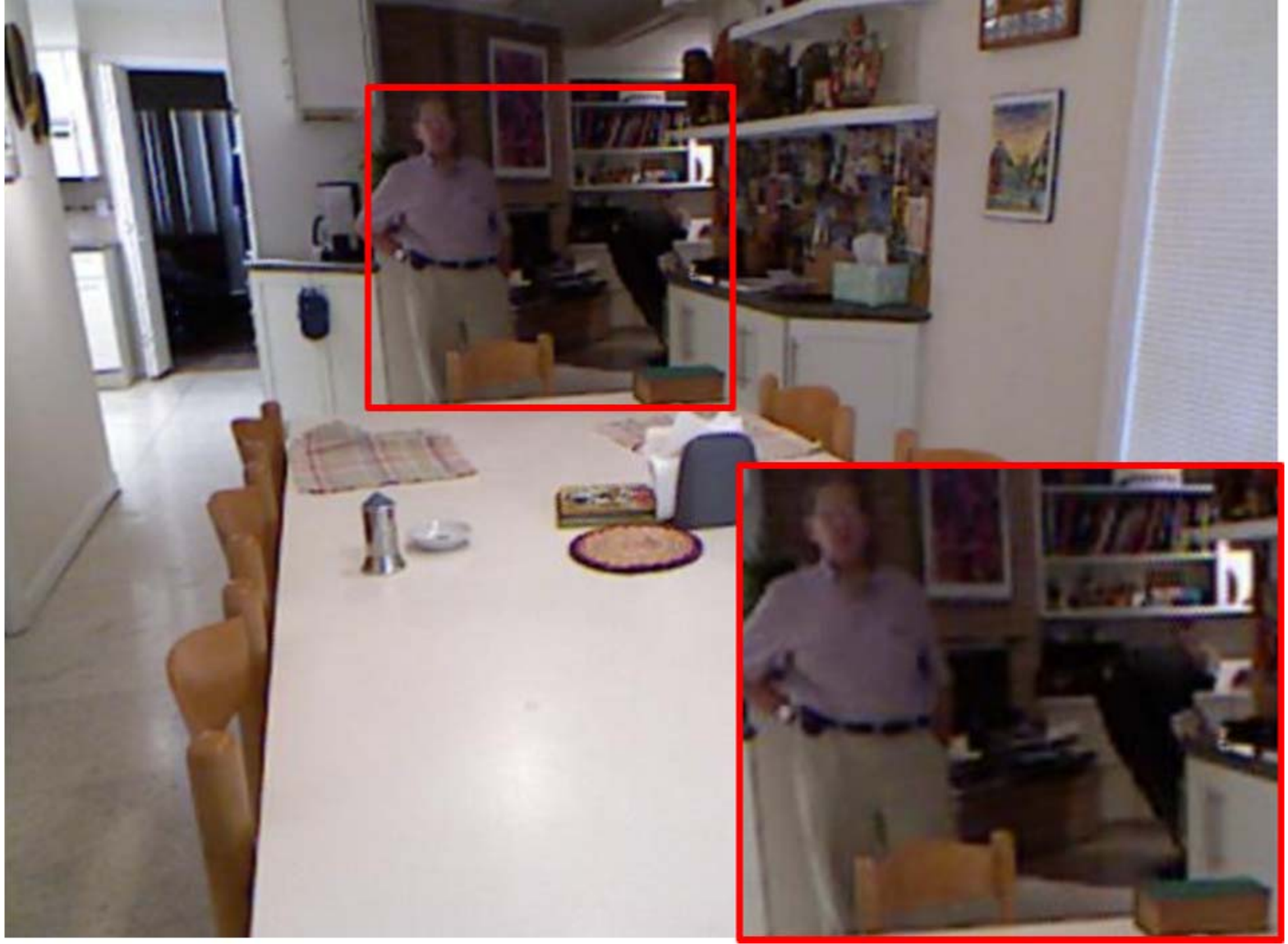}
         \\
       \hspace{-3mm}
         (a) Hazy input &
         (b) PFFNet~\cite{PFFNet}  &
         (c) MSBDN &
         (d) MSBDN-DFF &
         (e) Ground-truth
       \end{tabular}
     \end{adjustbox}
     \caption{
      \textbf{Visual results on the effect of the boosted decoder and dense feature fusion.}
     Compared with the baseline models, the proposed MSBDN-DFF model can generate a clearer image with more details.
     Best viewed on a high-resolution display.
     }
     \label{fig:visual_results_ablation}
     \vspace{-3mm}
     \end{figure*}
  

    \begin{table}[!t]
     \centering
     \caption{\textbf{Detection results on the KITTI Haze dataset.}
     We apply dehazing methods trained on the RESIDE dataset~\cite{RESIDE} to restore clean images and evaluate their perceptual quality for the object detection task.
     %
     %
     The mAP is the abbreviation of mean average precision.
     {\color{red}\textbf{Red texts}} indicate the best detection precision.
     }
  
     \vspace{0mm}
     \begin{minipage}{.49\textwidth}
        \begin{adjustbox}{width=\linewidth}
           \begin{tabu}{ccccccc}
           \tabucline[.5pt]{}
           \multicolumn{2}{c}{\textbf{YOLOv3}} \hspace{8pt} &Hazy  & DCP~\cite{He_dark}  & PFFNet~\cite{PFFNet} & DuRN~\cite{DuRN} & \multicolumn{1}{c}{Ours}  \\
           \hline
           \multirow{2}[2]{*}[2pt]{KITTI} & PSNR &10.35  &13.53 & 11.86 & 16.95 & {\color{red}\textbf{17.97}}  \\
                 & mAP &0.084  &0.239    &0.143   & 0.360  &{\color{red}\textbf{0.374}}  \\
           \hline
           \end{tabu}%
        \end{adjustbox}
        \end{minipage}
     \label{tab:heavy_kitti}%
     \vspace{-3mm}
  \end{table}

  \begin{table}[!t]
  \centering
  \caption{\textbf{Effect of the number of feature levels and ResBlocks.}
  $L$ denotes the number of feature levels and $B$ denotes the number of ResBlocks~\cite{resnet} in $G_{Res}$.
  All the experiments are conducted on the SOTS dataset~\cite{RESIDE}.
  {\color{red}\textbf{Red texts}} indicate the best performance.
  }
  \begin{minipage}{.45\textwidth}
   \begin{adjustbox}{width=\linewidth}
      \begin{tabu}{lcccc}
      \tabucline[.5pt]{}
      \multicolumn{2}{c}{\textbf{Configurations}} \hspace{8pt} &PFFNet~\cite{PFFNet}  & MSBDN  & MSBDN-DFF \\
      \hline
      \multirow{2}[2]{*}[2pt]{$L=4$, $B=8$} & PSNR &28.23  &30.92  &{\color{red}\textbf{32.07}}  \\
            & Param &2.6M  &3.1M   &4.5M   \\
      \hline
      \multirow{2}[2]{*}[2pt]{$L=5$, $B=8$} & PSNR &28.60   &32.00  &{\color{red}\textbf{32.93}}  \\
      &Param &10.2M  &12.6M   &19.6M   \\
      \hline
      \multirow{2}[2]{*}[2pt]{$L=5$, $B=18$} & PSNR &29.22  &32.85  &{\color{red}\textbf{33.79}}  \\
      &Param &22M  &24M   &31M   \\
      \hline
      \end{tabu}%
   \end{adjustbox}
   \end{minipage}
  \label{tab:configurations}%
  \vspace{-6mm}
  \end{table}

  \noindent{\bf Study of the network configuration.}
  To investigate the effect of the network configuration,
  we evaluate the PFFNet~\cite{PFFNet}, MSBDN, and MSBDN-DFF on the RESIDE dataset~\cite{RESIDE} under different network configurations.
  The PFFNet~\cite{PFFNet} adopts an encoder-decoder architecture without boosted modules and can be served as a baseline of our methods.
  The MSBDN is the encoder-decoder architecture with the SOS boosted modules, and the MSBDN-DFF is the proposed method with both the SOS boosted modules and DFF modules.
  
  Since all the above-mentioned methods are built upon the encoder-decoder architecture, we study the effect of two network parameters:
  the number of feature levels $L$ and the number of ResBlocks $B$ in the feature restoration module $G_{Res}$.
  The quantitative results are shown in Table~\ref{tab:configurations}.
  As the proposed SOS boosted modules and DFF modules can effectively extract features, larger numbers of levels~($L$) and ResBlocks~($B$) would lead to higher performance on our method.
  Moreover, introducing the SOS boosted modules and DFF modules can bring significant performance gains under different configurations,
  which demonstrates the effectiveness of the boosting and back-projection algorithms for image dehazing.
  For all the other experiments, we use $L=5$ and $B=18$ as the default network configurations.

  \begin{table*}[!t]
    \large
  \footnotesize
    \centering
    \caption{\textbf{Analysis on each component of the MSBDN-DFF.}
    All the methods are evaluated on the SOTS dataset \cite{RESIDE} using the same training setting as the proposed algorithm.
    {\color{red}\textbf{Red texts}} indicate the best performance of each part.
    }
    \begin{adjustbox}{width=\linewidth}
      \begin{tabu}{rcccccc|ccccc}
      \tabucline[1pt]{1-12}
      \multirow{2}[4]{*}[5pt]{\textbf{Baselines}} & \multicolumn{6}{c|}{\textbf{Effectiveness of SOS boosting strategy}} & \multicolumn{5}{c}{\textbf{Effectiveness of the dense feature fusion}} \\
  & PFFNet & Diffusion & Twicing  & Pyramid &U-Net & MSBDN & MSBDN-S & MSBDN-M  & MSBDN+ &PFFNet-DFF & MSBDN-DFF \\
      \hline
      boosting algorithm &       & \checkmark  & \checkmark  & \checkmark  & \checkmark    &\checkmark       & \checkmark      & \checkmark      & \checkmark  &    &\checkmark  \\
      FPN-like        &       &    &   & \checkmark   &  &       &       &       &    &   &  \\
      strengthened feature       &       &    &   &    &   & \checkmark     & \checkmark     & \checkmark     & \checkmark  &   & \checkmark \\
      dense feature fusion &       &    &   &    &   &       &\checkmark       & \checkmark     &   & \checkmark &\checkmark  \\
      simultaneously BP     &       &    &   &   &    &       &       & \checkmark      &    &  &  \\
      progressively BP  &       &    &   &    &   &       &       &       &     & \checkmark & \checkmark \\
      \hline
      Parameters & 22M   & 24M &24M  & 24M &24M  & 24M     & 26M   & 31M  & 31M &29M & 31M \\
      PSNR  & 29.22 & 27.45 & 32.49 & 31.53 &32.31 & {\color{red}\textbf{32.85}}  & 33.02 & 33.24 & 33.16 &32.95 & {\color{red}\textbf{33.79}} \\
      \tabucline[1pt]{}
      \end{tabu}%
    \end{adjustbox}
    \label{tab:2}%
  \end{table*}%

  \noindent{\bf Effectiveness of the SOS boosting strategy.}
  %
  The proposed MSBDN is based on an encoder-decoder architecture and the design of the boosted decoder is motivated by the SOS boosting strategy.
  We evaluate the following alternatives to demonstrate the effectiveness of the SOS boosting strategy.
  Starting from the baseline PFFNet model~\cite{PFFNet},
  we first remove the skip connections and add a diffusion module (shown in \figref{2}(a)) to each level of the decoder to implement the Diffusion Decoder.
  Next, we re-introduce the skip connections and change the diffusion modules with the twicing modules in~\figref{2}(b) and the pyramid modules in~\figref{2}(c)
  to construct the Twicing Decoder and Pyramid Decoder.
  Finally, we also evaluate the original U-Net Decoder in~\figref{2}(d) for a thorough study.

  The evaluation results are shown in Table~\ref{tab:2}.
  The network with the simple Diffusion Decoder performs the worst as expected.
  In general, the networks with the boosting modules achieve significant performance improvements over the PFFNet without using boosting strategies.
  Furthermore, the MSBDN with the proposed boosted decoder outperforms other boosting strategies by a margin of 0.36 dB at least.
  %
  We note that the MSBDN model improves the performance without introducing any extra layers,
  which demonstrates that the SOS boosting strategy better fits the problem and can benefit image dehazing algorithms.

  \figref{visual_results_ablation} shows an example of the visual results.
  The MSBDN model with the SOS boosted decoder performs better in haze removal.
  %
  Visualizations of the learned features are provided in the supplementary material.

  \noindent{\bf Effectiveness of the dense feature fusion.}
  The DFF module is designed to remedy the spatial information from high-resolution features and exploit the non-adjacent features.
  To demonstrate the effectiveness of the proposed DFF module, we evaluate several alternative solutions to the DFF.

  Starting from the MSBDN, we first fuse all the preceding features using the sampling operator and the bottleneck layer, as stated in the first paragraph of \secref{3.2}.
  This straightforward fusion strategy is referred to as the MSBDN-S.
  The results in~\tabref{2} show that the average PSNR value of the dehazed images by
  the MSBDN-S is 0.17 dB higher than that of the MSBDN,
  which demonstrate the benefits of exploiting the non-adjacent preceding features.
  %
  %
  To extract useful information from preceding features, we construct the MSBDN-M model by incorporating the back-projection technique~\cite{Irani1991}, as an alternative to the proposed DFF, into the MSBDN.
  %
  %
  In the MSBDN-M model, the reconstruction errors of all the preceding features are mapped back to the estimated feature simultaneously.
  On the contrary, the proposed DFF module adopts a progressive process to fuse one preceding feature at a time.
  Since extra layers are introduced by the DFF module,
  we construct an enhanced MSBDN~(referred to as the MSBDN+) by adding two residual blocks into each residual group for fair comparisons.
  The MSBDN+ model has similar parameters as the MSBDN-M and MSBDN-DFF schemes.
  The results show that the back-projection algorithm~\cite{Irani1991} in the MSBDN-M model is less effective with a small improvement margin over the MSBDN+ scheme.
  It is noted that the MSBDN-DFF model outperforms the MSBDN-M scheme by a margin of 0.55 dB without introducing any extra layers,
  which shows the effectiveness of the proposed DFF modules.
  In addition, we apply the proposed DFF module to the PFFNet (referred to as the PFFNet-DFF).
  As shown in \tabref{2}, the PFFNet-DFF achieves 3.73 dB performance improvement over the PFFNet,
  which demonstrates that the proposed DFF module can be easily deployed into other multi-scale architectures to improve the performance.
  
  As shown in~\figref{visual_results_ablation}(d),
  by remedying the spatial information and exploiting the preceding features, the MSBDN-DFF successfully removes the remaining haze in~\figref{visual_results_ablation}(c) and recovers more details.

  \section{Conclusions}
  We propose an effective Multi-Scale Boosted Dehazing Network with Dense Feature Fusion for image dehazing.
  The MSBDN is constructed on an encoder-decoder architecture, where the boosted decoder is designed based on the SOS boosting strategy.
  The DFF module is designed on the back-projection scheme, which can preserve the spatial information and exploit the features from non-adjacent levels.
  The ablation studies demonstrate that the proposed modules are effective for the dehazing problem.
  Extensive evaluations show that the proposed model performs favorably against state-of-the-art methods on the image dehazing datasets.
  %


\begin{thebibliography}{10}\itemsep=-1pt

\bibitem{NTIRE2018}
Cosmin Ancuti, Codruta~O Ancuti, and Radu Timofte.
\newblock Ntire 2018 challenge on image dehazing: Methods and results.
\newblock In {\em IEEE Conference on Computer Vision and Pattern Recognition
  Workshops}, pages 891--901, 2018.

\bibitem{IHAZE}
Cosmin Ancuti, Codruta~O Ancuti, Radu Timofte, and Christophe De~Vleeschouwer.
\newblock I-haze: a dehazing benchmark with real hazy and haze-free indoor
  images.
\newblock In {\em International Conference on Advanced Concepts for Intelligent
  Vision Systems}, pages 620--631. Springer, 2018.

\bibitem{OHAZE}
Codruta~O Ancuti, Cosmin Ancuti, Radu Timofte, and Christophe De~Vleeschouwer.
\newblock O-haze: a dehazing benchmark with real hazy and haze-free outdoor
  images.
\newblock In {\em IEEE Conference on Computer Vision and Pattern Recognition
  Workshops}, pages 754--762, 2018.

\bibitem{NLD}
Dana Berman, Shai Avidan, et~al.
\newblock Non-local image dehazing.
\newblock In {\em IEEE Conference on Computer Vision and Pattern Recognition},
  pages 1674--1682, 2016.

\bibitem{deep_physical1}
Bolun Cai, Xiangmin Xu, Kui Jia, Chunmei Qing, and Dacheng Tao.
\newblock Dehazenet: An end-to-end system for single image haze removal.
\newblock {\em IEEE Transactions on Image Processing}, 25(11):5187--5198, 2016.

\bibitem{TwicingBoosting}
Michael Charest, Michael Elad, and Peyman Milanfar.
\newblock A general iterative regularization framework for image denoising.
\newblock In {\em IEEE Conference on Information Sciences and Systems}, pages
  452--457, 2006.

\bibitem{DBF}
Chang Chen, Zhiwei Xiong, Xinmei Tian, and Feng Wu.
\newblock Deep boosting for image denoising.
\newblock In {\em European Conference on Computer Vision}, pages 3--18, 2018.

\bibitem{DBF_PAMI}
Chang Chen, Zhiwei Xiong, Xinmei Tian, Zheng-Jun Zha, and Feng Wu.
\newblock Real-world image denoising with deep boosting.
\newblock {\em IEEE Transactions on Pattern Analysis and Machine Intelligence},
  2019.

\bibitem{GCANet}
Dongdong Chen, Mingming He, Qingnan Fan, Jing Liao, Liheng Zhang, Dongdong Hou,
  Lu Yuan, and Gang Hua.
\newblock Gated context aggregation network for image dehazing and deraining.
\newblock In {\em IEEE Winter Conference on Applications of Computer Vision},
  pages 1375--1383, 2019.

\bibitem{DPN}
Yunpeng Chen, Jianan Li, Huaxin Xiao, Xiaojie Jin, Shuicheng Yan, and Jiashi
  Feng.
\newblock Dual path networks.
\newblock In {\em Neural Information Processing Systems}, pages 4467--4475,
  2017.

\bibitem{FPN_Pose}
Yilun Chen, Zhicheng Wang, Yuxiang Peng, Zhiqiang Zhang, Gang Yu, and Jian Sun.
\newblock Cascaded pyramid network for multi-person pose estimation.
\newblock In {\em IEEE Conference on Computer Vision and Pattern Recognition},
  pages 7103--7112, 2018.

\bibitem{BiBP}
Shengyang Dai, Mei Han, Ying Wu, and Yihong Gong.
\newblock Bilateral back-projection for single image super resolution.
\newblock In {\em IEEE International Conference on Multimedia and Expo}, pages
  1039--1042, 2007.

\bibitem{RI-GAN}
Akshay Dudhane, Harshjeet Singh~Aulakh, and Subrahmanyam Murala.
\newblock Ri-gan: An end-to-end network for single image haze removal.
\newblock In {\em IEEE Conference on Computer Vision and Pattern Recognition
  Workshops}, pages 1--10, 2019.

\bibitem{cycledehaze}
Deniz Engin, Anil Gen{\c{c}}, and Hazim Kemal~Ekenel.
\newblock Cycle-dehaze: Enhanced cyclegan for single image dehazing.
\newblock In {\em IEEE Conference on Computer Vision and Pattern Recognition},
  pages 825--833, 2018.

\bibitem{tr_dehaze1}
Raanan Fattal.
\newblock Single image dehazing.
\newblock {\em ACM Transactions on Graphics}, 27(3):72--92, 2008.

\bibitem{tr_dehaze3}
Raanan Fattal.
\newblock Dehazing using color-lines.
\newblock {\em ACM Transactions on Graphics}, 34(1):13--31, 2014.

\bibitem{residual_grid}
Damien Fourure, R{\'e}mi Emonet, Elisa Fromont, Damien Muselet, Alain Tremeau,
  and Christian Wolf.
\newblock Residual conv-deconv grid network for semantic segmentation.
\newblock In {\em British Machine Vision Conference}, 2017.

\bibitem{D-DIP}
Yossi Gandelsman, Assaf Shocher, and Michal Irani.
\newblock " double-dip": Unsupervised image decomposition via coupled
  deep-image-priors.
\newblock In {\em IEEE Conference on Computer Vision and Pattern Recognition},
  pages 11026--11035, 2019.

\bibitem{kitti_detection}
Andreas Geiger, Philip Lenz, and Raquel Urtasun.
\newblock Are we ready for autonomous driving? the kitti vision benchmark
  suite.
\newblock In {\em IEEE Conference on Computer Vision and Pattern Recognition},
  pages 3354--3361, 2012.

\bibitem{monodepth2}
Clément Godard, Oisin Mac~Aodha, Michael Firman, and Gabriel~J. Brostow.
\newblock Digging into self-supervised monocular depth prediction.
\newblock In {\em IEEE International Conference on Computer Vision}, pages
  3828--3838, 2019.

\bibitem{IBCNN}
Shizhong Han, Zibo Meng, Ahmed-Shehab Khan, and Yan Tong.
\newblock Incremental boosting convolutional neural network for facial action
  unit recognition.
\newblock In {\em Neural Information Processing Systems}, pages 109--117, 2016.

\bibitem{DBPN}
Muhammad Haris, Greg Shakhnarovich, and Norimichi Ukita.
\newblock Deep back-projection networks for super-resolution.
\newblock In {\em IEEE Conference on Computer Vision and Pattern Recognition},
  pages 1664--1673, 2018.

\bibitem{He_dark}
Kaiming He, Jian Sun, and Xiaoou Tang.
\newblock Single image haze removal using dark channel prior.
\newblock {\em IEEE Transactions on Pattern Analysis and Machine Intelligence},
  33(12):2341--2353, 2011.

\bibitem{resnet}
Kaiming He, Xiangyu Zhang, Shaoqing Ren, and Jian Sun.
\newblock Deep residual learning for image recognition.
\newblock In {\em IEEE Conference on Computer Vision and Pattern Recognition},
  pages 770--778, 2016.

\bibitem{MsDense}
Gao Huang, Danlu Chen, Tianhong Li, Felix Wu, Laurens van~der Maaten, and
  Kilian~Q Weinberger.
\newblock Multi-scale dense networks for resource efficient image
  classification.
\newblock In {\em International Conference on Learning Representations}, 2018.

\bibitem{desnenet}
Gao Huang, Zhuang Liu, Laurens Van Der~Maaten, and Kilian~Q Weinberger.
\newblock Densely connected convolutional networks.
\newblock In {\em IEEE Conference on Computer Vision and Pattern Recognition},
  pages 4700--4708, 2017.

\bibitem{Irani1991}
Michal Irani and Shmuel Peleg.
\newblock Improving resolution by image registration.
\newblock {\em Cvgip Graphical Models and Image Processing}, 53(3):231--239,
  1991.

\bibitem{adam}
Diederik~P Kingma and Jimmy Ba.
\newblock Adam: A method for stochastic optimization.
\newblock In {\em International Conference on Learning Representations}, 2015.

\bibitem{PFPN}
Alexander Kirillov, Ross Girshick, Kaiming He, and Piotr Doll{\'a}r.
\newblock Panoptic feature pyramid networks.
\newblock In {\em IEEE Conference on Computer Vision and Pattern Recognition},
  pages 6399--6408, 2019.

\bibitem{LapSRN}
Wei-Sheng Lai, Jia-Bin Huang, Narendra Ahuja, and Ming-Hsuan Yang.
\newblock Deep laplacian pyramid networks for fast and accurate
  super-resolution.
\newblock In {\em IEEE Conference on Computer Vision and Pattern Recognition},
  pages 624--632, 2017.

\bibitem{AOD}
Boyi Li, Xiulian Peng, Zhangyang Wang, Jizheng Xu, and Dan Feng.
\newblock Aod-net: All-in-one dehazing network.
\newblock In {\em IEEE International Conference on Computer Vision}, pages
  4770--4778, 2017.

\bibitem{dehazing_app2}
Boyi Li, Xiulian Peng, Zhangyang Wang, Jizheng Xu, and Dan Feng.
\newblock End-to-end united video dehazing and detection.
\newblock In {\em Thirty-Second AAAI Conference on Artificial Intelligence},
  pages 7016--7023, 2018.

\bibitem{RESIDE}
Boyi Li, Wenqi Ren, Dengpan Fu, Dacheng Tao, Dan Feng, Wenjun Zeng, and
  Zhangyang Wang.
\newblock Reside: A benchmark for single image dehazing.
\newblock {\em IEEE Transactions on Image Processing}, 28(1):492--505, 2018.

\bibitem{li2018image}
Jinjiang Li, Guihui Li, and Hui Fan.
\newblock Image dehazing using residual-based deep cnn.
\newblock {\em IEEE Access}, 6:26831--26842, 2018.

\bibitem{DcGAN}
Runde Li, Jinshan Pan, Zechao Li, and Jinhui Tang.
\newblock Single image dehazing via conditional generative adversarial network.
\newblock In {\em IEEE Conference on Computer Vision and Pattern Recognition},
  pages 8202--8211, 2018.

\bibitem{VHaze}
Zhuwen Li, Ping Tan, Robby~T. Tan, Danping Zou, Steven~Zhiying Zhou, and
  Loong-Fah Cheong.
\newblock Simultaneous video defogging and stereo reconstruction.
\newblock In {\em IEEE Conference on Computer Vision and Pattern Recognition},
  pages 4988--4997, 2015.

\bibitem{FSSD}
Zuoxin Li and Fuqiang Zhou.
\newblock Fssd: feature fusion single shot multibox detector.
\newblock {\em arXiv}, 2017.

\bibitem{EDSR}
Bee Lim, Sanghyun Son, Heewon Kim, Seungjun Nah, and Kyoung~Mu Lee.
\newblock Enhanced deep residual networks for single image super-resolution.
\newblock In {\em IEEE Conference on Computer Vision and Pattern Recognition
  Workshops}, pages 1132--1140, 2017.

\bibitem{FPN}
Tsung-Yi Lin, Piotr Doll{\'a}r, Ross Girshick, Kaiming He, Bharath Hariharan,
  and Serge Belongie.
\newblock Feature pyramid networks for object detection.
\newblock In {\em IEEE Conference on Computer Vision and Pattern Recognition},
  pages 2117--2125, 2017.

\bibitem{griddehazenet}
Xiaohong Liu, Yongrui Ma, Zhihao Shi, and Jun Chen.
\newblock Griddehazenet: Attention-based multi-scale network for image
  dehazing.
\newblock In {\em IEEE International Conference on Computer Vision}, pages
  7314--7323, 2019.

\bibitem{DuRN}
Xing Liu, Masanori Suganuma, Zhun Sun, and Takayuki Okatani.
\newblock Dual residual networks leveraging the potential of paired operations
  for image restoration.
\newblock In {\em IEEE Conference on Computer Vision and Pattern Recognition},
  pages 7007--7016, 2019.

\bibitem{DeepPriorsDehaze}
Yang Liu, Jinshan Pan, Jimmy Ren, and Zhixun Su.
\newblock Learning deep priors for image dehazing.
\newblock In {\em IEEE International Conference on Computer Vision}, pages
  2492--2500, 2019.

\bibitem{PFFNet}
Kangfu Mei, Aiwen Jiang, Juncheng Li, and Mingwen Wang.
\newblock Progressive feature fusion network for realistic image dehazing.
\newblock In {\em Asian Conference on Computer Vision}, pages 203--215, 2018.

\bibitem{DiffusionBoosting}
Peyman Milanfar.
\newblock A tour of modern image filtering: New insights and methods, both
  practical and theoretical.
\newblock {\em IEEE Signal Processing Magazine}, 30(1):106--128, 2013.

\bibitem{BoostCNN}
Mohammad Moghimi, Serge~J Belongie, Mohammad~J Saberian, Jian Yang, Nuno
  Vasconcelos, and Li-Jia Li.
\newblock Boosted convolutional neural networks.
\newblock In {\em British Machine Vision Conference}, pages 24.1--24.13, 2016.

\bibitem{jinshan_dual}
Jinshan Pan, Sifei Liu, Deqing Sun, Jiawei Zhang, Yang Liu, Jimmy Ren, Zechao
  Li, Jinhui Tang, Huchuan Lu, Yu-Wing Tai, et~al.
\newblock Learning dual convolutional neural networks for low-level vision.
\newblock In {\em IEEE Conference on Computer Vision and Pattern Recognition},
  pages 3070--3079, 2018.

\bibitem{Pix2pixHaze}
Yanyun Qu, Yizi Chen, Jingying Huang, and Yuan Xie.
\newblock Enhanced pix2pix dehazing network.
\newblock In {\em IEEE Conference on Computer Vision and Pattern Recognition},
  pages 8160--8168, 2019.

\bibitem{yolov3}
Joseph Redmon and Ali Farhadi.
\newblock Yolov3: an incremental improvement.
\newblock {\em arXiv}, 2018.

\bibitem{MSCNN}
Wenqi Ren, Si Liu, Hua Zhang, Jinshan Pan, Xiaochun Cao, and Ming-Hsuan Yang.
\newblock Single image dehazing via multi-scale convolutional neural networks.
\newblock In {\em European Conference on Computer Vision}, pages 154--169,
  2016.

\bibitem{GFN}
Wenqi Ren, Lin Ma, Jiawei Zhang, Jinshan Pan, Xiaochun Cao, Wei Liu, and
  Ming-Hsuan Yang.
\newblock Gated fusion network for single image dehazing.
\newblock In {\em IEEE Conference on Computer Vision and Pattern Recognition},
  pages 3253--3261, 2018.

\bibitem{SOS}
Yaniv Romano and Michael Elad.
\newblock Boosting of image denoising algorithms.
\newblock {\em Siam Journal on Imaging Sciences}, 8(2):1187--1219, 2015.

\bibitem{UNet}
Olaf Ronneberger, Philipp Fischer, and Thomas Brox.
\newblock U-net: Convolutional networks for biomedical image segmentation.
\newblock In {\em International Conference on Medical Image Computing and
  Computer-assisted Intervention}, pages 234--241, 2015.

\bibitem{dehazing_app1}
Christos Sakaridis, Dengxin Dai, and Luc Van~Gool.
\newblock Semantic foggy scene understanding with synthetic data.
\newblock {\em International Journal of Computer Vision}, pages 1--20, 2018.

\bibitem{Middlebury}
Daniel Scharstein and Richard Szeliski.
\newblock High-accuracy stereo depth maps using structured light.
\newblock In {\em IEEE Conference on Computer Vision and Pattern Recognition},
  pages 195--202, 2003.

\bibitem{NYU}
Nathan Silberman, Derek Hoiem, Pushmeet Kohli, and Rob Fergus.
\newblock Indoor segmentation and support inference from rgbd images.
\newblock In {\em European Conference on Computer Vision}, pages 746--760,
  2012.

\bibitem{tr_dehaze2}
Robby~T Tan.
\newblock Visibility in bad weather from a single image.
\newblock In {\em IEEE Conference on Computer Vision and Pattern Recognition},
  pages 1--8, 2008.

\bibitem{SSIM}
Zhou Wang, Alan~C Bovik, Hamid~R Sheikh, Eero~P Simoncelli, et~al.
\newblock Image quality assessment: from error visibility to structural
  similarity.
\newblock {\em IEEE Transactions on Image Processing}, 13(4):600--612, 2004.

\bibitem{PDN}
Dong Yang and Jian Sun.
\newblock Proximal dehaze-net: A prior learning-based deep network for single
  image dehazing.
\newblock In {\em European Conference on Computer Vision}, pages 702--717,
  2018.

\bibitem{deep_physical_unsupervised1}
Xitong Yang, Zheng Xu, and Jiebo Luo.
\newblock Towards perceptual image dehazing by physics-based disentanglement
  and adversarial training.
\newblock In {\em Thirty-second AAAI conference on Artificial Intelligence},
  pages 7485--7492, 2018.

\bibitem{DCPDN}
He Zhang and Vishal~M Patel.
\newblock Densely connected pyramid dehazing network.
\newblock In {\em IEEE Conference on Computer Vision and Pattern Recognition},
  pages 3194--3203, 2018.

\bibitem{MsPPN}
He Zhang, Vishwanath Sindagi, and Vishal~M Patel.
\newblock Multi-scale single image dehazing using perceptual pyramid deep
  network.
\newblock In {\em IEEE Conference on Computer Vision and Pattern Recognition
  Workshops}, pages 902--911, 2018.

\bibitem{GFN_BMVC}
Xinyi Zhang, Hang Dong, Zhe Hu, Wei-Sheng Lai, Fei Wang, and Ming-Hsuan Yang.
\newblock Gated fusion network for joint image deblurring and super-resolution.
\newblock In {\em British Machine Vision Conference}, pages 153--165, 2018.

\bibitem{HazeRD}
Yanfu Zhang, Li Ding, and Gaurav Sharma.
\newblock Hazerd: an outdoor scene dataset and benchmark for single image
  dehazing.
\newblock In {\em IEEE International Conference on Image Processing}, pages
  3205--3209, 2017.

\bibitem{RDN}
Yulun Zhang, Yapeng Tian, Yu Kong, Bineng Zhong, and Yun Fu.
\newblock Residual dense network for image super-resolution.
\newblock In {\em IEEE Conference on Computer Vision and Pattern Recognition},
  pages 2472--2481, 2018.

\bibitem{error_grid}
Jonathan Zung, Ignacio Tartavull, Kisuk Lee, and H~Sebastian Seung.
\newblock An error detection and correction framework for connectomics.
\newblock In {\em Neural Information Processing Systems}, pages 6818--6829,
  2017.

\end{thebibliography}


\small

\bibliographystyle{ieee_fullname}

\end{document}